\documentclass{article}

\usepackage[nonatbib,final]{neurips_2022}

\usepackage[utf8]{inputenc} 
\usepackage[T1]{fontenc}    
\usepackage[breaklinks,colorlinks]{hyperref}       
\usepackage{url}            
\usepackage{booktabs}       
\usepackage{amsfonts}       
\usepackage{nicefrac}       
\usepackage{microtype}      
\usepackage{xcolor}         
\usepackage{overpic}

\usepackage{graphicx}
\usepackage{amsmath}
\usepackage{amssymb}
\usepackage{cite}
\usepackage{multirow}
\usepackage{caption}
\usepackage{makecell}
\usepackage{bbding} 
\usepackage{lipsum}
\usepackage{bm}
\usepackage[american]{babel}

\usepackage{subfloat}
\usepackage{float}
\usepackage{subfig}

\usepackage{tabularx}
\usepackage{adjustbox}
\usepackage{wrapfig}

\RequirePackage{xspace}

\makeatletter
\DeclareRobustCommand\onedot{\futurelet\@let@token\@onedot}
\def\@onedot{\ifx\@let@token.\else.\null\fi\xspace}

\def\eg{\emph{e.g}\onedot} 
\def\ie{\emph{i.e}\onedot} 
 
\def\etc{\emph{etc}\onedot}

\def\etal{\emph{et al}\onedot}
\makeatother

\usepackage[capitalize]{cleveref}
\crefname{section}{Sec.}{Secs.}
\Crefname{section}{Section}{Sections}
\Crefname{table}{Table}{Tables}
\crefname{table}{Tab.}{Tabs.}
\Crefname{equation}{Equation}{Equations}
\crefname{equation}{Eqn.}{Eqns.}

\DeclareMathOperator*{\argmin}{\arg\min}

\newcommand{\tabincell}[2]{\begin{tabular}{@{}#1@{}}#2\end{tabular}}

\title{Self-Supervised Image Restoration\\ with Blurry and Noisy Pairs}

\author{%
  Zhilu Zhang$^1$, Rongjian Xu$^{1}$, Ming Liu$^{1}$, Zifei Yan$^{1,}$\thanks{Correspondence author.} , Wangmeng Zuo$^{1,2}$\\
   $^1$ Harbin Institute of Technology, Harbin, China; \\ 
   $^2$ Peng Cheng Laboratory, China \\
  \texttt{cszlzhang@outlook.com, ronjon.xu@gmail.com,} \\
  \texttt{csmliu@outlook.com, yanzifei@hit.edu.cn, wmzuo@hit.edu.cn}
}

\begin{document}

\maketitle

\begin{abstract}
When taking photos under an environment with insufficient light, the exposure time and the sensor gain usually require to be carefully chosen to obtain images with satisfying visual quality. For example, the images with high ISO usually have inescapable noise, while the long-exposure ones may be blurry due to camera shake or object motion. Existing solutions generally suggest to seek a balance between noise and blur, and learn denoising or deblurring models under either full- or self-supervision. However, the real-world training pairs are difficult to collect, and the self-supervised methods merely rely on blurry or noisy images are limited in performance. In this work, we tackle this problem by jointly leveraging the short-exposure noisy image and the long-exposure blurry image for better image restoration. Such setting is practically feasible due to that short-exposure and long-exposure images can be either acquired by two individual cameras or synthesized by a long burst of images. Moreover, the short-exposure images are hardly blurry, and the long-exposure ones have negligible noise. Their complementarity makes it feasible to learn restoration model in a self-supervised manner. Specifically, the noisy images can be used as the supervision information for deblurring, while the sharp areas in the blurry images can be utilized as the auxiliary supervision information for self-supervised denoising. By learning in a collaborative manner, the deblurring and denoising tasks in our method can benefit each other. Experiments on synthetic and real-world images show the effectiveness and practicality of the proposed method. Codes are available at \url{https://github.com/cszhilu1998/SelfIR}.
\end{abstract}

\section{Introduction}
It is a common yet challenging task to acquire visually appealing photos with appropriate brightness under a low light environment.
Traditional ways to increase the image brightness include enlarging the aperture, adopting a higher ISO, and reducing the shutter speed (\ie, lengthening exposure time).
As for smartphone cameras with fixed aperture, brightness can only be adjusted by setting the sensor gain (\ie, ISO) and exposure time. 
Nonetheless, they are negatively correlated to maintain the appropriate brightness level of the image, \ie, the shorter-exposure image generally adopts a higher ISO, while the longer-exposure image usually has a lower ISO.
Moreover, high ISO configuration introduces inevitable and complex noise due to the limited photon amount and the process of camera image signal processing (ISP) pipeline, while long-exposure is prone to produce blurry images due to the camera shake and scene variations.
Consequently, photographers have to make a compromise between noise and blur.

Recent advances in image restoration make it possible to further improve the visual quality of the acquired low light images by leveraging deep image denoising or deblurring networks.
Taking supervised denoising as an example, 
synthetic or real-world noisy-clean image pairs are required to train the deep networks~\cite{DnCNN,FFDNet,MWCNN,CBDNet,DANet,MIRNet,Restormer}.
However, the models trained with synthetic training pairs are hard to generalize to real noisy images, and the real-world clean reference images are usually obtained by averaging hundreds of noisy ones~\cite{SIDD} or with complicated capturing and processing procedure~\cite{DND}, making the collection of large scale real-world training datasets laborious, expensive, and time-consuming.
Such problems greatly limit the deployment of models on more devices with different noise distributions.
Alternatively, a surge of self-supervised image denoising methods~\cite{DIP,Noise2Noise,Nei2Nei,N2V,Laine19,DBSN,Blind2Unblind,AP-BSN,R2R,soltanayev2018training,zhussip2019extending,CVF-SID,IDR} have been developed to avoid the collection of ground-truth (GT) training images, yet are limited in handling complex real-world image noise.
Another possible solution is to perform motion deblurring on long-exposure images.
Early explorations~\cite{fergus2006removing,shan2008high,xu2013unnatural,michaeli2014blind,schuler2015learning,ren2018deep} are mainly given to  spatially uniform deblurring caused by camera motion.
Recently, the proposal of non-uniform datasets (\eg, GoPro~\cite{DeepDeblur}, REDS~\cite{nah2019ntire}, HIDE~\cite{HIDE}, RealBlur~\cite{RealBlur}, \etc) has greatly boosted the research of deblurring in more practical scenes involving both camera shake and object motion~\cite{DeepDeblur,zhang2018dynamic,tao2018scale,MIMO-UNet,MPRNet}.

In this paper, we suggest to improve low light imaging by jointly leveraging the short-exposure noisy and long-exposure blurry images.
First, such setting is practically feasible. 
For example, multiple cameras have been equipped in modern smartphones, which can be designed to acquire short-exposure and long-exposure images, respectively.
Moreover, one can also synthesize a pair of blurry and noisy images from a long burst of images captured by a camera. 
Second, the noisy and blurry images convey complementary information, which is beneficial to improve restoration performance and makes self-supervised image restoration (SelfIR) possible.
We note that several methods~\cite{SIGGRAPH-2007,LSD2,Low-light-TMM,ks2022content} have been suggested to combine the blurry image with their noisy counterpart for better image restoration, yet it remains 
uninvestigated under the self-supervised regime.

We further present a SelfIR model with blurry and noisy pairs.
Even though the blurry and noisy images are both \textit{disturbed}, the short-exposure images taken with high ISO are \textit{hardly blurry}, while the long-exposure images taken with low ISO are generally near \textit{noise-free}.
Thus, the long-exposure and short-exposure images can be used to provide some supervision information for each other.
On the one hand, the noisy image can serve as an alternative of sharp image to supervise deblurring with negligible performance degradation.
On the other hand, the static regions in long-exposure images are noise-free and sharp, which in turn can provide auxiliary supervision information for image denoising.
Taking these two aspects into account, we present a collaborative learning (co-learning) method termed SelfIR for deblurring and denoising, which is effective in leveraging the complementary information of long- and short-exposure images and can be learned in a self-supervised manner.

Extensive experiments on synthetic data are conducted to evaluate our SelfIR.
Both quantitative and qualitative results show that SelfIR outperforms the state-of-the-art self-supervised denoising methods, as well as the supervised denoising and deblurring counterparts.
To further verify the practicality of our SelfIR model, we have also collected a set of 61 real-world blurry and noisy pairs using smartphones.
Since there are no corresponding ground-truth images for calculating full-reference image quality assessment (IQA) metrics, we evaluate the restoration results using no-reference IQA metrics.
The results show that our method also performs favorably against the competing methods.

To sum up, the main contributions of this work include:
\begin{itemize}
    \item We take a step forward in leveraging blurry and noisy image pairs for image restoration. Going beyond leveraging their complementarity in improving restoration performance, we show that it can also be utilized for self-supervised learning of the restoration model.
    \item A self-supervised image restoration model (SelfIR) is proposed, where short-exposure images serve as supervision for the corresponding deblurring task, while the sharp regions in long-exposure images provide auxiliary supervision for self-supervised denoising. 
    \item Extensive experiments on both synthetic and real-world image pairs show that our SelfIR performs favorably against the state-of-the-art self-supervised denoising methods, as well as the baseline supervised deblurring and denoising methods.
\end{itemize}

\clearpage

\section{Related Work}\label{sec:related_work}
In this section, we briefly review burst image denoising and deblurring, as well as self-supervised image denoising and deblurring methods.
In addition, we recommend \cite{delbracio2021mobile} for a comprehensive introduction to the relevant mobile computational photography.

\noindent\textbf{Burst Image Denoising and Deblurring.}
In comparison with a single image, burst images can provide more information that is beneficial for image restoration.
Hasinoff~\etal~\cite{HDRplus} utilize an FFT-based alignment algorithm and a hybrid 2D/3D Wiener filter to denoise and merge a burst of underexposed frames for low-light photography.
KPN~\cite{KPN} predicts spatially variant kernels for every burst noisy image to merge them. 
BPN~\cite{BPN} extends the KPN method with a basis prediction network and achieves larger denoising kernels under certain computing resource constraints.
Aittala~\etal~\cite{aittala2018burst} take both noise and blur into account, and restore sharp and noise-free images from burst images in an order-independent manner.

\noindent\textbf{Self-Supervised Image Denoising and Deblurring.}
Recently, self-supervised learning has drawn upsurging attention in low-level vision. 
DIP~\cite{DIP} utilizes the image prior implicitly captured by the network structure to repair corrupted images.
SelfDeblur~\cite{SelfDeblur} respectively models the deep priors of clear image and blur kernel for self-supervised deblurring.
For these methods, the networks are required to re-train from scratch for each test image, which is less efficient, especially for mobile or edge devices.
Noise2Noise~\cite{Noise2Noise} demonstrates that noisy pairs with mutually independent noise can be used to train a denoising network, opening the door to self-supervised denoising.
Neighbor2Neighbor~\cite{Nei2Nei} utilizes a random neighbor sub-sampler to generate the training pairs from noisy images themselves.
In addition, some works~\cite{N2V,Laine19,DBSN,Blind2Unblind,AP-BSN} elaborately design blind-spot networks to avoid learning the identity mapping for self-supervised denoising.
However, the self-supervised denoising methods are limited in handling complex image noise.
In this work, we utilize the complementarity of long-exposure blurry and short-exposure noisy images for better self-supervised image restoration.

\section{Proposed Method}\label{sec:proposed_method}
In this section, we first show the feasibility of taking noisy images as the supervision of deblurring.
Then, we introduce the sharp area detection method in long-exposure images and auxiliary loss for self-supervised denoising.
Finally, we present the proposed co-learning framework SelfIR.
\subsection{Deblurring with Noisy Image}\label{sec:3.1}
When taking long-exposure photos, the shake of the camera and the motion of objects usually lead to a blurry image $\mathbf{I}_\mathcal{B}$, which can be formulated by,
\begin{equation}
    \mathbf{I}_\mathcal{B} = \mathcal{K}(\mathbf{I})+\mathbf{N}_\mathcal{B},
    \label{eqn:blur}
\end{equation}
where $\mathbf{I}$ is the latent clear image, $\mathcal{K}$ denotes the blur process with non-uniform kernels, $\mathbf{N}_\mathcal{B}$ represents the low-intensity noise.
Existing supervised deblurring methods generally utilize a deep neural network (denoted by $\mathcal{D_B}$) to estimate $\mathbf{I}$ from $\mathbf{I}_\mathcal{B}$.
For training the parameters of $\mathcal{D_B}$, which is denoted by $\Theta_\mathcal{B}$, the optimization objective can be defined by,
\begin{equation}
    \Theta_\mathcal{B}^\ast = \arg\min_{\Theta_\mathcal{B}} \mathbb{E}_{\mathbf{I}_\mathcal{B},\mathbf{I}} \left[\mathcal{L}\left(\mathcal{D_B}(\mathbf{I}_\mathcal{B}; \Theta_\mathcal{B}), \mathbf{I} \right)\right],
    \label{eqn:deblur}
\end{equation}
where $\mathcal{L}$ denotes the loss functions for supervised learning.
However, collecting clear images is troublesome in real-world scenes.
Inspired by Noise2Noise~\cite{Noise2Noise}, we show that the noisy short-exposure image can be a substitution of the latent clear image to supervise the task of deblurring.

When taking short-exposure photos under low light environment, the limited photon amount and inherent defects of camera ISP make the images noisy (denoted by $\mathbf{I}_\mathcal{N}$), which can be formulated by, 
\begin{equation}
    \mathbf{I}_\mathcal{N} = \mathbf{I}+\mathbf{N}_\mathcal{N},
    \label{eqn:noise}
\end{equation}
where $\mathbf{N}_\mathcal{N}$ represents the noise (with much higher intensity than $\mathbf{N}_\mathcal{B}$).
When using the noisy image $\mathbf{I}_\mathcal{N}$ as the supervision of deblurring, the optimization of $\Theta_\mathcal{B}$ can be expressed as,
\begin{equation}
\small
  \Theta_\mathcal{B}^\ast = \arg\min_{\Theta_\mathcal{B}} \mathbb{E}_{\mathbf{I}_\mathcal{B},\mathbf{I}_\mathcal{N}} \left[\mathcal{L}\left(\mathcal{D_B}(\mathbf{I}_\mathcal{B}; \Theta_\mathcal{B}), \mathbf{I}_\mathcal{N} \right)\right] = \arg\min_{\Theta_\mathcal{B}} \mathbb{E}_{\mathbf{I}_\mathcal{B}} \left[\mathbb{E}_{\mathbf{I}_\mathcal{N}|\mathbf{I}_\mathcal{B}} \left[\mathcal{L}\left(\mathcal{D_B}(\mathbf{I}_\mathcal{B}; \Theta_\mathcal{B}), \mathbf{I}_\mathcal{N} \right)\right]\right].
\label{eqn:deblur_un}
\end{equation}
Suppose that the loss function $\mathcal{L}$ in \cref{eqn:deblur_un} is the $\ell_2$ loss, we have,
\begin{equation}
\begin{split}
   \mathbb{E}_{\mathbf{I}_\mathcal{N}|\mathbf{I}_\mathcal{B}} \left[\mathcal{L}\left(\mathcal{D_B}(\mathbf{I}_\mathcal{B}; \Theta_\mathcal{B}), \mathbf{I}_\mathcal{N} \right)\right] 
   & = \mathbb{E}_{\mathbf{I}_\mathcal{N}|\mathbf{I}_\mathcal{B}} \left[\| \mathcal{D_B}(\mathbf{I}_\mathcal{B}; \Theta_\mathcal{B}) - \mathbf{I}_\mathcal{N} \|_2^2 \right] \\
   & = \mathbb{E}_{\mathbf{I}, \mathbf{N}_\mathcal{N}|\mathbf{I}_\mathcal{B}} \left[\| \mathcal{D_B}(\mathbf{I}_\mathcal{B}; \Theta_\mathcal{B}) - (\mathbf{I} + \mathbf{N}_\mathcal{N}) \|_2^2 \right] \\
   & = \mathbb{E}_{\mathbf{I}|\mathbf{I}_\mathcal{B}} \left[\| \mathcal{D_B}(\mathbf{I}_\mathcal{B}; \Theta_\mathcal{B}) - \mathbf{I} \|_2^2 \right] - \\
   & \quad \ \, 2 \mathbb{E}_{\mathbf{I}, \mathbf{N}_\mathcal{N}|\mathbf{I}_\mathcal{B}} \left[(\mathcal{D_B}(\mathbf{I}_\mathcal{B}; \Theta_\mathcal{B}) - \mathbf{I})^\top \mathbf{N}_\mathcal{N}\right] + \\
   & \quad \ \, \mathbb{E}_{\mathbf{N}_\mathcal{N}|\mathbf{I}_\mathcal{B}} \left[\| \mathbf{N}_\mathcal{N} \|_2^2 \right],
\end{split}
\label{eqn:deblur_un1}
\end{equation}
where $\mathbb{E}_{\mathbf{N}_\mathcal{N}|\mathbf{I}_\mathcal{B}} \left[\| \mathbf{N}_\mathcal{N} \|_2^2\right]$ can be regarded as a constant and be safely discarded from \cref{eqn:deblur_un1}.
Further, assume that $\mathbf{N}_\mathcal{N}$ is zero-mean, $\mathbf{N}_\mathcal{N}$ and $\mathbf{I}$ are independent, we can get,
\begin{equation}
    \mathbb{E}_{\mathbf{I}, \mathbf{N}_\mathcal{N}|\mathbf{I}_\mathcal{B}} \left[(\mathcal{D_B}(\mathbf{I}_\mathcal{B}; \Theta_\mathcal{B}) - \mathbf{I})^\top \mathbf{N}_\mathcal{N}\right]=0 .
\end{equation}
In this case, the optimal solution $\Theta_\mathcal{B}^\ast$ in \cref{eqn:deblur_un} and that in \cref{eqn:deblur} are the same.
Thus, it is feasible to utilize noisy short-exposure images instead of clear ones as the supervision of deblurring.

\subsection{Denoising with Long-Exposure Image}\label{sec:3.2}
{\parfillskip0pt\par}
\begin{wrapfigure}{R}{0.47\textwidth}
	\centering
	\includegraphics[width=0.97\linewidth]{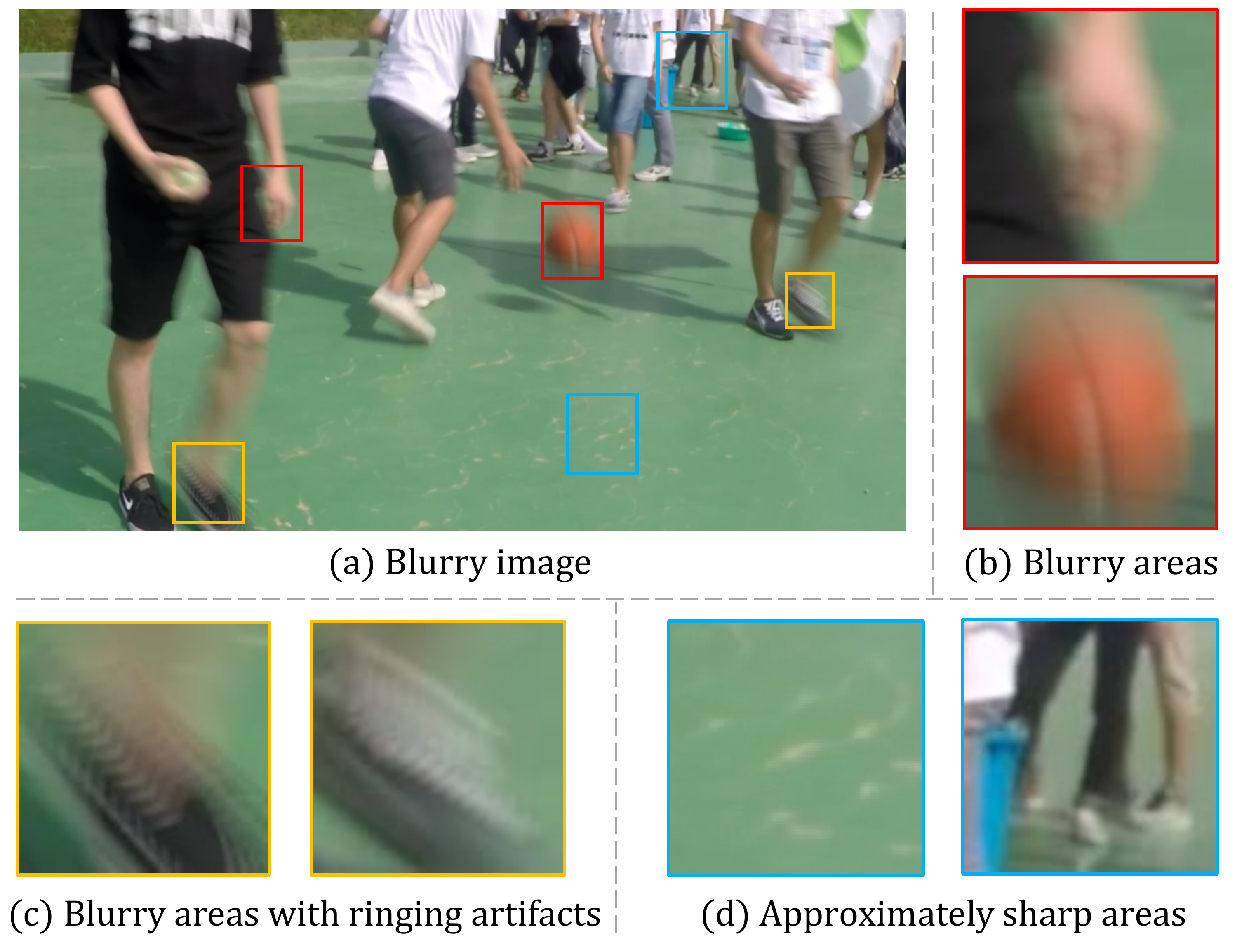}
	\caption{A blurry image example where the blur is non-uniform in (a). It includes some common blurry areas (b), severe blurry areas with ringing artifacts (c), and some approximately sharp areas (d).}
	\label{fig:blurry}
	\vspace{-2mm}
\end{wrapfigure}
%
%
Self-supervised denoising makes it possible to remove the noise without clean image supervision, but may give rise to obvious performance degradation, especially when handling complex real-world noises.
Here we propose to alleviate this problem by introducing some extra supervision from the long-exposure counterpart.
Obviously, taking the whole long-exposure image as supervision will bring adverse effects, making the results to be blurry.
Nonetheless, it is worth noting that, the blur process $\mathcal{K}$ in \cref{eqn:blur} is generally non-uniform, and sometimes not all areas are blurry.
As shown in \cref{fig:blurry}(d), there exist some approximately sharp regions in the long-exposure image, which can provide partial supervision information that benefits self-supervised denoising.

Therefore, it is crucial to pinpoint the sharp areas in the long-exposure image.
Otherwise, we would prefer to go without sharp areas than accept a shoddy option, as misjudgments of sharp areas will lead to worse denoising results.
However, without any discriminative clues, it is very likely to misjudge the sharp areas.
Considering that $\mathbf{I}_\mathcal{N}$ is nearly non-blurry due to the short exposure time, it may be a reference for sharp area detection.
To avoid noise interference, we pre-process $\mathbf{I}_\mathcal{N}$ with a self-supervised denoising model, and take the result $\mathbf{\tilde{I}}_\mathcal{N}$ to help detect sharp regions in the corresponding blurry image $\mathbf{I}_\mathcal{B}$.

Specifically, we first divide $\mathbf{I}_\mathcal{B}$ and $\mathbf{\tilde{I}}_\mathcal{N}$ into $\mathit{N}$ non-overlapping patches. 
For each patch pair $\mathbf{I}_\mathcal{B}^\mathit{n}$ and $\mathbf{\tilde{I}}_\mathcal{N}^\mathit{n}$ ($ 1 \leq \mathit{n} \leq \mathit{N} $), our goal is to obtain a mask $\mathit{m}^\mathit{n} \in \{0,1\}$ that indicates whether $\mathbf{I}_\mathcal{B}^\mathit{n}$ is a sharp patch.
Since $\mathbf{\tilde{I}}_\mathcal{N}$ is nearly non-blurry, when some severe motion blurs exist in $\mathbf{I}_\mathcal{B}^\mathit{n}$, the difference between $\mathbf{I}_\mathcal{B}^\mathit{n}$ and $\mathbf{\tilde{I}}_\mathcal{N}^\mathit{n}$ in textures and edges should be evident.
Taking the above into account, we adopt a similarity metric $\mathit{s}$ to detect the areas with severe motion blurs, \ie, 
\begin{equation}
    \mathit{m}^\mathit{n} = \mathtt{sgn}(\mathtt{max}(0, \mathit{s}(\mathbf{I}_\mathcal{B}^\mathit{n},\mathbf{\tilde{I}}_\mathcal{N}^\mathit{n}) - \epsilon_\mathit{s})),
    \label{mask_1}
\end{equation}
where structural similarity (SSIM)~\cite{SSIM} is utilized for the similarity metric $\mathit{s}$, $\epsilon_s$ denotes the threshold, while $\mathtt{max}(a,b)$ and $\mathtt{sgn}(\cdot)$ denote the maximum and sign function, respectively.

However, the initial denoising result $\mathbf{\tilde{I}}_\mathcal{N}$ may be over-smooth, in other words, \cref{mask_1} may fail when facing some mildly blurred regions in $\mathbf{I}_\mathcal{B}^\mathit{n}$.
Therefore, we further measure the difference in variance between $\mathbf{I}_\mathcal{B}^\mathit{n}$ and $\mathbf{\tilde{I}}_\mathcal{N}^\mathit{n}$.
When the variance of $\mathbf{I}_\mathcal{B}^\mathit{n}$ is greater than that of $\mathbf{\tilde{I}}_\mathcal{N}^\mathit{n}$, we consider that $\mathbf{I}_\mathcal{B}^\mathit{n}$ is potential to be a sharp patch.
It should be noted that the difference in variance is not suitable for detecting some severely blurry areas with ringing artifacts (see \cref{fig:blurry}(c)).
In such blurry regions, the variance of $\mathbf{I}_\mathcal{B}^\mathit{n}$ may also be greater than that of $\mathbf{\tilde{I}}_\mathcal{N}^\mathit{n}$.
When synthesizing blurry images, some works~\cite{nah2019ntire,wieschollek2017learning,zhou2019davanet} remove the artifacts by interpolating the frames before averaging the sharp images.
However, the artifacts also exist in real-world blurry images, especially in areas with flickering lights.
Therefore, we still take the ringing artifacts into consideration in this work.

As a result, we jointly use the SSIM and variance measure for judging sharp regions, $\mathit{m}^\mathit{n}$ can be formulated as,
\begin{equation}
    \mathit{m}^\mathit{n} = \mathtt{sgn}(\mathtt{max}(0, \mathit{s}(\mathbf{I}_\mathcal{B}^\mathit{n},\mathbf{\tilde{I}}_\mathcal{N}^\mathit{n}) - \epsilon_\mathit{s})) * \mathtt{sgn}(\mathtt{max}(0, \mathtt{var}(\mathbf{I}_\mathcal{B}^\mathit{n}) - \mathtt{var}(\mathbf{\tilde{I}}_\mathcal{N}^\mathit{n}) - \epsilon_\mathit{v})),
    \label{mask_2}
\end{equation}
where $\mathtt{var}( \cdot ) $ and $\epsilon_\mathit{v}$ denote the variance function and the threshold, respectively.
$\epsilon_s$ and $\epsilon_v$ are set to 0.99 and 1e-5 when color intensity values of $\mathbf{I}_\mathcal{B}^\mathit{n}$ and $\mathbf{\tilde{I}}_\mathcal{N}^\mathit{n}$ are normalized to $[0,1]$. 
Defining the final output image as $\mathbf{\hat{I}}$, the auxiliary loss function for denoising can be denoted as,
\begin{equation}
    \mathcal{L}_\mathit{aux}(\mathbf{\hat{I}}, \mathbf{I}_\mathcal{B}) = {\sum}_{\mathit{n}=1}^\mathit{N} \mathit{m}^\mathit{n} \| \mathbf{\hat{I}}^\mathit{n} - \mathbf{I}_\mathcal{B}^\mathit{n} \|_2^2 .
    \label{loss:aux}
\end{equation}

\subsection{Co-Learning of Deblurring and Denoising}\label{sec:3.3}
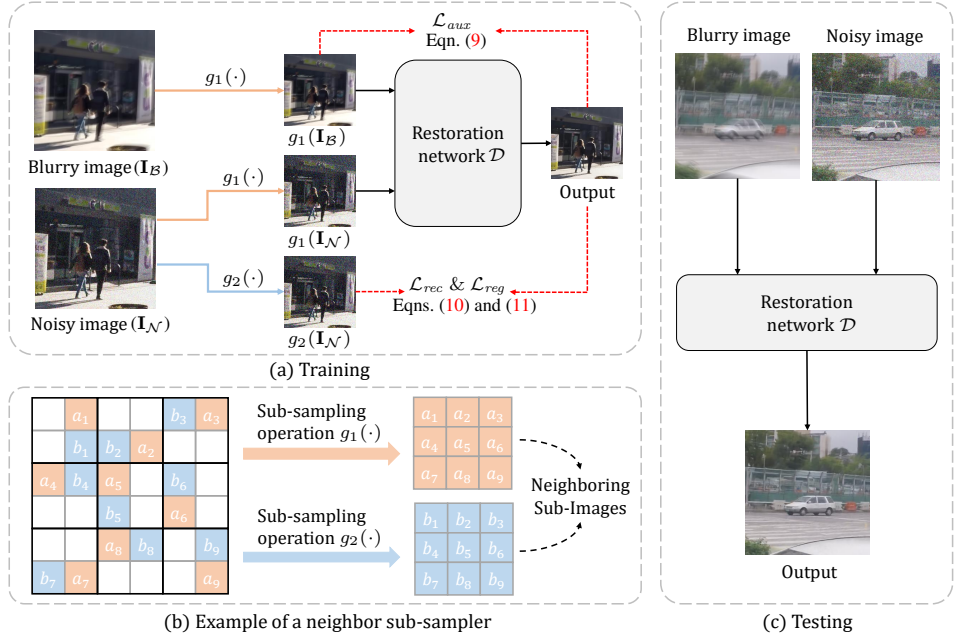
\begin{figure}[t]
	\centering
	\begin{overpic}[width=0.93\linewidth]{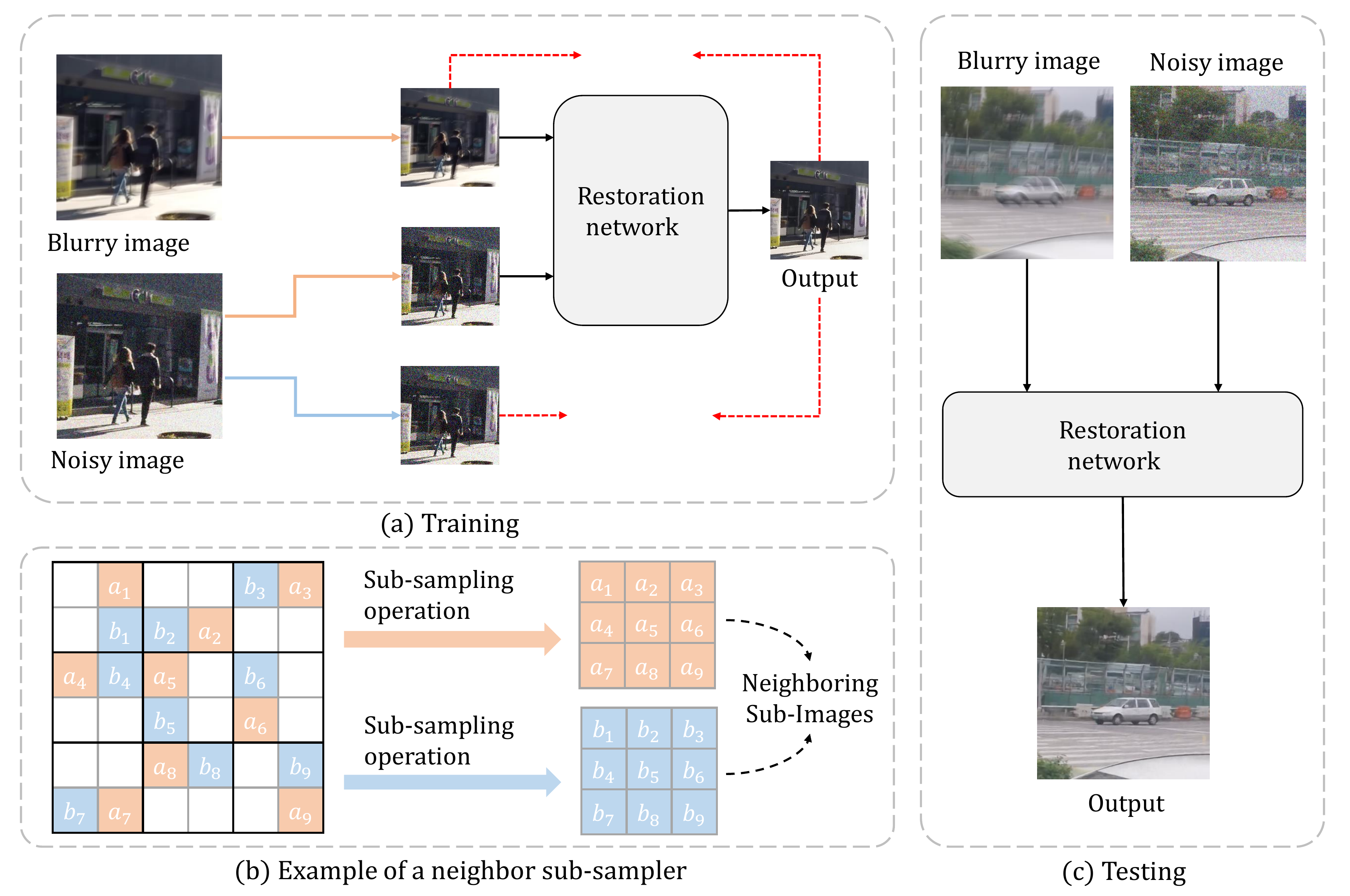}
	    \put(21.5,57.1){\scriptsize$\mathit{g}_1(\cdot)$}
	    \put(23.3,46.8){\scriptsize$\mathit{g}_1(\cdot)$}
	    \put(23.3,36.5){\scriptsize$\mathit{g}_2(\cdot)$}
	    \put(35.3,20.6){\scriptsize$\mathit{g}_1(\cdot)$}
	    \put(35.3,9.9){\scriptsize$\mathit{g}_2(\cdot)$}
	    
	    \put(14.3,47.8){\scriptsize($\mathbf{I}_\mathcal{B}$)}
	    \put(14,31.8){\scriptsize($\mathbf{I}_\mathcal{N}$)}
	    
	    \put(30,50.9){\scriptsize$\mathit{g}_1(\mathbf{I}_\mathcal{B})$}
	    \put(30,40.5){\scriptsize$\mathit{g}_1(\mathbf{I}_\mathcal{N})$}
	    \put(30,30.2){\scriptsize$\mathit{g}_2(\mathbf{I}_\mathcal{N})$}
	    
	    \put(44.8,62.8){\scriptsize$\mathcal{L}_\mathit{aux}$}
	    \put(44.2,60.8){\scriptsize\cref{loss:aux}}
	    \put(42.4,35.8){\scriptsize$\mathcal{L}_\mathit{rec}\ \&\ \mathcal{L}_\mathit{reg}$}
	    \put(40.8,33.6){\scriptsize\cref{loss:rec,loss:reg}}
	    
	    \put(50.6,49.0){\scriptsize$\mathcal{D}$}
	    \put(86.5,31.6){\scriptsize$\mathcal{D}$}
	    
	\end{overpic}
	\vspace{-1mm}
	\caption{Overview of our proposed SelfIR framework.
	(a) Training phase of SelfIR. Sub-sampled blurry image $\mathit{g}_1(\mathbf{I}_\mathcal{B})$ and noisy image $\mathit{g}_1(\mathbf{I}_\mathcal{N})$ are taken as the inputs. $\mathit{g}_2(\mathbf{I}_\mathcal{N})$ is used for calculating the reconstruction loss $\mathcal{L}_\mathit{rec}$ (see \cref{loss:rec}) and regularization loss $\mathcal{L}_\mathit{reg}$ (see \cref{loss:reg}), while $\mathit{g}_1(\mathbf{I}_\mathcal{B})$ is taken for calculating auxiliary loss (see \cref{loss:aux}).
	(b) Example of neighbor sub-sampler. In each $2\times2$ cell, two pixels are randomly selected for respectively composing the neighboring sub-images.
	(c) Testing phase of SelfIR. The blurry and noisy images can be directly taken for restoration.}
	\label{fig:framework}
	\vspace{-2mm}
\end{figure}

As illustrated in \cref{sec:3.1,sec:3.2}, the noisy images can provide effective supervision information for deblurring, while the blurry images can also provide auxiliary supervision for self-supervised denoising.
In other words, we can learn to deblur or denoise without additional clear and clean ground-truths.
However, learning them independently will lead to limited performance.
As some works~\cite{SIGGRAPH-2007,LSD2,Low-light-TMM,ks2022content} have suggested, better restoration performance can be obtained by incorporating blurry with noisy images.
Thus, it should be further considered to achieve self-supervised image restoration that takes two disturbed images as input.
In this section, we delicately design the self-supervised model SelfIR that learns from both deblurring and denoising tasks in a collaborative learning manner.

For training SelfIR, it is required to avoid the trivial solution when taking both blurry and noisy images as the input. 
For example, when taking noisy images for supervising deblurring, the model may simply output the noisy images instead of learning the deburring results.
Fortunately, Neighbor2Neighbor~\cite{Nei2Nei} shows the noises in two sub-sampled images (\ie, $\mathit{g}_1(\mathbf{I}_\mathcal{N})$ and $\mathit{g}_2(\mathbf{I}_\mathcal{N})$, see \cref{fig:framework}(b)) from noisy image $\mathbf{I}_\mathcal{N}$ are almost independent.
Therefore, they can be respectively taken as the input and target for training the denoising subtask in a self-supervised manner.
Simultaneously, combining the derivation in \cref{eqn:deblur_un1}, $\mathit{g}_1(\mathbf{I}_\mathcal{B})$ and $\mathit{g}_2(\mathbf{I}_\mathcal{N})$ can be respectively taken as the input and target to train the deblurring subtask.
Therefore, our SelfIR can take a step forward and deliver both $\mathit{g}_1(\mathbf{I}_\mathcal{B})$ and $\mathit{g}_1(\mathbf{I}_\mathcal{N})$ ($i.e.$, $\{\mathit{g}_1(\mathbf{I}_\mathcal{B}, \mathit{g}_1(\mathbf{I}_\mathcal{N})\}$) into the restoration network $\mathcal{D}$, as shown in \cref{fig:framework}(a).
The sub-sampled image $\mathit{g}_2(\mathbf{I}_\mathcal{N})$ is taken for calculating reconstruction loss $\mathcal{L}_\mathit{rec}$, which can be written as,
\begin{equation}
    \mathcal{L}_\mathit{rec} = \| \mathcal{D}(\mathit{g}_1(\mathbf{I}_\mathcal{B}), \mathit{g}_1(\mathbf{I}_\mathcal{N})) -  \mathit{g}_2(\mathbf{I}_\mathcal{N}) \|_2^2.
    \label{loss:rec}
\end{equation}
We also calculate regularization loss $\mathcal{L}_\mathit{reg}$ similar to \cite{Nei2Nei},
\begin{equation}
     \mathcal{L}_\mathit{reg} = \| \mathcal{D}(\mathit{g}_1(\mathbf{I}_\mathcal{B}), \mathit{g}_1(\mathbf{I}_\mathcal{N})) -  \mathit{g}_2(\mathbf{I}_\mathcal{N}) - ( \mathit{g}_1(\hat{\mathcal{D}}(\mathbf{I}_\mathcal{B}, \mathbf{I}_\mathcal{N})) - \mathit{g}_2(\hat{\mathcal{D}}(\mathbf{I}_\mathcal{B}, \mathbf{I}_\mathcal{N})) ) \|_2^2,
    \label{loss:reg}
\end{equation}
where $\hat{\mathcal{D}}$ has same parameters with $\mathcal{D}$ but has no gradient for back-propagation.
Moreover, the sharp areas in $\mathit{g}_1(\mathbf{I}_\mathcal{B})$ can  provide auxiliary supervision for the model learning and the performance can be further improved.
Specifically, we calculate auxiliary loss  $\mathcal{L}_\mathit{aux}(\mathcal{D}(\mathit{g}_1(\mathbf{I}_\mathcal{B}), \mathit{g}_1(\mathbf{I}_\mathcal{N}))), \mathit{g}_1(\mathbf{I}_\mathcal{B}))$ in \cref{loss:aux}.
For obtaining the mask $\mathit{m}^\mathit{n}$, we replace $\mathbf{I}_\mathcal{B}$ and $\mathbf{\tilde{I}}_\mathcal{N}$ in~\cref{mask_2} with $\mathit{g}_1(\mathbf{I}_\mathcal{B})$ and $\hat{\mathcal{D}}(\mathit{g}_1(\mathbf{I}_\mathcal{B}), \mathit{g}_1(\mathbf{I}_\mathcal{N}))$, respectively.
Finally, the learning objective of parameters $\Theta_\mathcal{D}$ with the co-learning manner can be formulated as,
\begin{equation}
  \Theta_\mathcal{D}^\ast = \argmin_{\Theta_\mathcal{D}} \mathbb{E}_{\mathbf{I}_\mathcal{B},\mathbf{I}_\mathcal{N}} ( \mathcal{L}_\mathit{rec} + \lambda_\mathit{reg} \mathcal{L}_\mathit{reg} + \lambda_\mathit{aux} \mathcal{L}_\mathit{aux}),
\label{eqn:co-learning}
\end{equation}
where $\lambda_\mathit{reg}$ and $\lambda_\mathit{aux}$  are hyper-parameters for balancing the loss terms.

\section{Experiments}\label{sec:experiments}
In this section, we first describe the dataset configuration and the training / evaluation protocols in detail.
Then, both the quantitative and qualitative results on synthetic and real-world images are given for comprehensive evaluation.
\subsection{Implementation Details}\label{sec:training_details}

\noindent\textbf{Synthetic Datasets.}
For synthesizing blurry images, early methods~\cite{SIGGRAPH-2007,LSD2} convolve sharp images with simulated blur kernels, which is quite different from the real-world blur model.
Recently, the GoPro dataset\footnote{\url{https://seungjunnah.github.io/Datasets/gopro.html}}~\cite{DeepDeblur} offers a more realistic way to synthesize blurry images, which has been widely adopted for motion deblurring tasks~\cite{DeepDeblur,zhang2018dynamic,tao2018scale,MIMO-UNet,MPRNet}.
In the dataset, although slight inherent noise exists in the sharp video frames that are captured by a high-speed camera (\ie, GoPro Hero4 Black), it has little effect on model training and evaluation.
We directly regard the sharp images as clean images.
The blurry image is generated by averaging consecutive sharp frames, which can better simulate both camera shake and object motion.
%
%
Thus, we can get the blurry-clear pairs as $\mathbf{I}_\mathcal{B}$ and $\mathbf{I}$, respectively.
For synthesizing the noisy image $\mathbf{I}_\mathcal{N}$, we consider three noise distributions: 1)~Gaussian noise with $\sigma\!\in\![5/255, 50/255]$, 2)~Poisson noise with $\lambda\!\in\![5, 50]$, and 3)~sensor noise~\cite{UPI}.
Please refer to the supplementary material for the noise formulations.
Experiments are conducted in both sRGB and raw-RGB spaces.
For the sRGB space, we add Gaussian or Poisson noise to the sRGB sharp images.
While for the raw-RGB space, the more complex sensor noise is added to the sharp raw-RGB images, and the blurry images are also converted back to the raw-RGB space through the unprocessing~\cite{UPI} pipeline.
Finally, there are 2,103 image pairs for training, and we use the remaining 1,111 pairs for testing.

\noindent\textbf{Real-world Datasets.}
Furthermore, we have also collected 61 real-world blurry and noisy raw-RGB pairs with Huawei P40 smartphones in the professional camera mode.
In particular, the blurry images are captured with a low ISO and long exposure time, while the corresponding noisy images are captured with a high ISO and short exposure time in the same scene.
Please refer to the supplementary material for detailed collection process.
Due to the limited amount of real-world images, we use 30 pairs to fine-tune the model pre-trained on raw-RGB images with synthetic sensor noise, and take the remaining 31 pairs for evaluation.

\noindent\textbf{Training Details.}
Following~\cite{Noise2Noise,Laine19,Nei2Nei}, we adopt a U-Net~\cite{UNet} architecture as our restoration network, where two encoders are respectively deployed to the blurry and noisy input images for better domain-specific feature extraction, and then blurry and noisy features are fused in the decoder part.
Detailed architecture is given in the supplementary material.
During training, the batch size is set to 16 and the patch size is $128 \times 128$.
Adam optimizer~\cite{Adam} with $\beta_1=0.9$ and $\beta_2=0.999$ is used to train the network for 200 epochs.
The learning rate is initially set to $3\times10^{-4}$ for synthetic experiments and $1\times10^{-4}$ for real-world experiments.
And it reduces by half every 50 epochs.
For the hyper-parameters in~\cref{eqn:co-learning}, $\lambda_\mathit{aux}$ is set to 2, $\lambda_\mathit{reg}$ is set to 2 and 4 for experiments in sRGB space and raw-RGB space, respectively.
All experiments are conducted with PyTorch~\cite{PyTorch} on an Nvidia GeForce RTX 2080Ti GPU.

\noindent\textbf{Evaluation Configurations.}
We convert all raw-RGB results to sRGB space through post-processing pipeline, and the quantitative metrics are computed in the sRGB space.
For the results of synthetic experiments, we take peak signal to noise ratio (PSNR), structural similarity (SSIM)~\cite{SSIM} and learned perceptual image patch similarity (LPIPS)~\cite{LPIPS} as evaluation metrics.
For the results of real-world experiments, due to the lack of ground-truth images, we utilize no-reference IQA metrics (\ie, NIQE~\cite{NIQE}, NRQM~\cite{NRQM}, and PI~\cite{PI}) to evaluate the generated images.

\begin{table}[t]
  \small
  \renewcommand\arraystretch{1}
  \begin{center}
    \caption{Quantitative results on synthetic sRGB images.}
    \label{table:sRGB}
    \vspace{2mm}
    \scalebox{1}{
    \begin{tabular}{cccccc}
      \toprule
      & Method
      & \tabincell{c}{Gaussian $\sigma \in [5/255,50/255]$ \\ \footnotesize{PSNR}\scriptsize{$\uparrow$} \small{/} \footnotesize{SSIM}\scriptsize{$\uparrow$} \small{/} \footnotesize{LPIPS}\scriptsize{$\downarrow$}}
      & \tabincell{c}{Poisson $\lambda \in [5,50]$  \\ \footnotesize{PSNR}\scriptsize{$\uparrow$} \small{/} \footnotesize{SSIM}\scriptsize{$\uparrow$} \small{/} \footnotesize{LPIPS}\scriptsize{$\downarrow$}} \\

      \midrule
      \multirow{2}{*}{\begin{tabular}[c]{@{}c@{}}Supervised\\Deblurring \end{tabular}}
      & Baseline$_\mathcal{B}$  & \multicolumn{2}{c}{28.24 / 0.8561 / 0.191} \\
      & DeepDeblur~\cite{DeepDeblur}  & \multicolumn{2}{c}{30.04 / 0.9015 / 0.133} \\

      \midrule
      \multirow{2}{*}{\begin{tabular}[c]{@{}c@{}}Supervised\\Denoising \end{tabular}}
      & Baseline$_\mathcal{N}$  & 34.91 / 0.9360 / 0.098  & 33.15 / 0.9225 /  0.126 \\
      & DnCNN~\cite{DnCNN}  & 34.63 / 0.9308 / 0.121 & 32.45 / 0.9084 / 0.128 \\

      \midrule
      Supervised IR & Baseline$_\mathcal{R}$       & 36.15 / 0.9534 / 0.070 & 34.74 / 0.9454 / 0.084 \\

      \midrule
      \multirow{8}{*}{\begin{tabular}[c]{@{}c@{}}Self-Supervised\\Denoising \end{tabular}}
      & N2N~\cite{Noise2Noise}            & 34.88 / 0.9354 / 0.100 & 33.09 / 0.9216 / 0.129 \\
      & N2V~\cite{N2V}           & 33.09 / 0.9180 / 0.115 & 31.81 / 0.8999 / 0.137 \\
      & Laine19-mu~\cite{Laine19}         & 33.61 / 0.9227 / 0.104 & 32.29 / 0.9091 / 0.131 \\
      & Laine19-pme~\cite{Laine19}         & 34.76 / 0.9322 / 0.086 & 32.77 / 0.9147 / 0.116 \\
      & DBSN~\cite{DBSN}        & 33.72 / 0.9224 / 0.111 & 31.46 / 0.8883 / 0.144 \\
      & R2R~\cite{R2R}        & 33.74 / 0.9223 / 0.100 & 30.05 / 0.7649 / 0.230 \\
      & Neighbor2Neighbor~\cite{Nei2Nei}  & 34.29 / 0.9271 / 0.085 & 32.68 / 0.9160 / 0.111 \\
      & Blind2Unblind~\cite{Blind2Unblind}  & 34.69 / 0.9353 / 0.107 & 33.09 / 0.9216 / 0.132 \\
      
      \midrule
      \multirow{1}{*}{Ours}
      & SelfIR          & 35.74 / 0.9499 / 0.076 & 34.27 / 0.9404 / 0.092 \\
      \bottomrule
    \end{tabular}}
  \end{center}
\end{table}

\subsection{Experimental Results in sRGB Space}
To assess our proposed SelfIR, we build several baseline methods for comparison, including 1)~a supervised deblurring baseline (denoted by Baseline$_\mathcal{B}$), 2)~a supervised denoising baseline (denoted by Baseline$_\mathcal{N}$), 3)~a supervised image restoration baseline which takes both blurry and noisy images as input (denoted by Baseline$_\mathcal{R}$), and 4)~Neighbor2Neighbor~\cite{Nei2Nei} since the denoising part of our co-learning framework is based on it.
Note that the network structure is not the focus of this paper, so we deploy a simple U-Net~\cite{UNet} architecture for our restoration network $\mathcal{D}$ and the first three baseline methods.
Besides, we choose two classical supervised methods (\ie, DeepDeblur~\cite{DeepDeblur} and DnCNN~\cite{DnCNN}) for comparison.
Self-supervised denoising methods (\eg, N2N~\cite{Noise2Noise}, N2V~\cite{N2V}, Laine19~\cite{Laine19}, DBSN~\cite{DBSN}, R2R~\cite{R2R}, and Blind2Unblind~\cite{Blind2Unblind}) are also compared.
For DeepDeblur~\cite{DeepDeblur}, we use the officially released model for testing.
For other methods, the models are retrained with our synthetic images for a fair comparison.

The quantitative results of synthetic experiments on Gaussian and Poisson noise are given in \cref{table:sRGB}.
One can see that our SelfIR outperforms all other methods except the supervised IR model (Baseline$_\mathcal{R}$), which can be seen as the upper limit of the restoration performance with the utilized U-Net architecture and the blurry-noisy pair input.
In comparison to the self-supervised denoising baseline Neighbor2Neighbor~\cite{Nei2Nei}, 1.45 dB and 1.59 dB PSNR gains are obtained by our SelfIR on Gaussian and Poisson noise, respectively.
Even compared to the supervised denoising baseline model (Baseline$_\mathcal{N}$), our method also improves the PSNR index by 0.83 dB and 1.12 dB.
Due to the severely ill-posed nature of deblurring, SelfIR brings a PSNR gain by more than 6 dB comparing to the supervised deblurring baseline model (Baseline$_\mathcal{B}$).
The results clearly show that the blurry images can indeed provide beneficial features for denoising, and so do the noisy images for deblurring.

The qualitative results are shown in \cref{fig:gau1,fig:poi1}.
One can see that our results are sharper and clearer than other denoising or deblurring methods.
In terms of visual quality, the proposed SelfIR is almost comparable to the supervised image restoration model (Baseline$_\mathcal{R}$), and combines the advantages of supervised denoising Baseline$_\mathcal{N}$ and deblurring Baseline$_\mathcal{B}$.
More results can be found in the supplemental material.

 \begin{figure}[t]
    \vspace{-2mm}
    \begin{minipage}[t]{\linewidth}
        \centering
        \tiny
        \subfloat[\scriptsize Noisy Image]
        {
            \includegraphics[height=.15\linewidth, width=.15\linewidth]{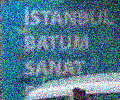}
        }
        \hfill
        \subfloat[\scriptsize Baseline$_\mathcal{N}$]
        {
            \includegraphics[height=.15\linewidth, width=.15\linewidth]{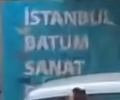}
        }
        \hfill
        \subfloat[\scriptsize N2V~\cite{N2V}]
        {
            \includegraphics[height=.15\linewidth, width=.15\linewidth]{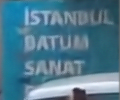}
        }
        \hfill
        \subfloat[\scriptsize Laine19-pme~\cite{Laine19}]
        {
            \includegraphics[height=.15\linewidth, width=.15\linewidth]{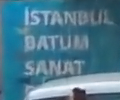}
        }
        \hfill
        \subfloat[\scriptsize DBSN~\cite{DBSN}]
        {
            \includegraphics[height=.15\linewidth, width=.15\linewidth]{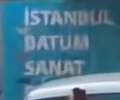}
        }
        \hfill
        \subfloat[\scriptsize R2R~\cite{R2R}]
        {
            \includegraphics[height=.15\linewidth, width=.15\linewidth]{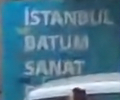}
        }
        \vspace{-2mm}

        \subfloat[\scriptsize Blurry Image]
        {
            \includegraphics[height=.15\linewidth, width=.15\linewidth]{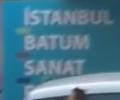}
        }
        \hfill
        \subfloat[\scriptsize Baseline$_\mathcal{B}$]
        {
            \includegraphics[height=.15\linewidth, width=.15\linewidth]{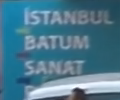}
        }
        \hfill        
        \subfloat[\tiny Neighbor2Neighbor~\cite{Nei2Nei}]
        {
            \includegraphics[height=.15\linewidth, width=.15\linewidth]{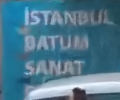}
        }
        \hfill
        \subfloat[\scriptsize Blind2Unblind~\cite{Blind2Unblind}]
        {
            \includegraphics[height=.15\linewidth, width=.15\linewidth]{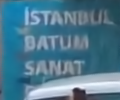}
        }
        \hfill
        \subfloat[\scriptsize SelfIR (Ours)]
        {
            \includegraphics[height=.15\linewidth, width=.15\linewidth]{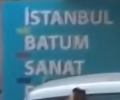}
        }
        \hfill
        \subfloat[\scriptsize Baseline$_\mathcal{R}$]
        {
            \includegraphics[height=.15\linewidth, width=.15\linewidth]{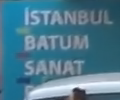}
        }
    \end{minipage}
    \vspace{-1mm}
    \caption{Visual comparison on Gaussian noise. The texts in our result are sharper and clearer. The proposed SelfIR is almost comparable to supervised Baseline$_\mathcal{R}$, which can be seen as the upper limit of the restoration performance with the utilized U-Net architecture and the blurry-noisy pair input.}
    \label{fig:gau1}
    \vspace{-3mm}
\end{figure}

\begin{figure}[t]
    \vspace{-2mm}
    \begin{minipage}[t]{\linewidth}
        \centering
        \tiny
        \subfloat[\scriptsize Noisy Image]
        {
            \includegraphics[height=.15\linewidth, width=.15\linewidth]{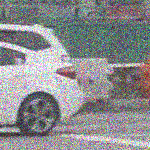}
        }
        \hfill
        \subfloat[\scriptsize Baseline$_\mathcal{N}$]
        {
            \includegraphics[height=.15\linewidth, width=.15\linewidth]{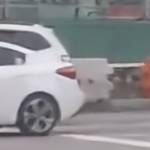}
        }
        \hfill
        \subfloat[\scriptsize N2V~\cite{N2V}]
        {
            \includegraphics[height=.15\linewidth, width=.15\linewidth]{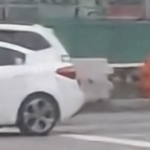}
        }
        \hfill
        \subfloat[\scriptsize Laine19-pme~\cite{Laine19}]
        {
            \includegraphics[height=.15\linewidth, width=.15\linewidth]{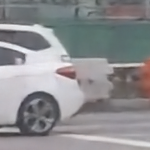}
        }
        \hfill
        \subfloat[\scriptsize DBSN~\cite{DBSN}]
        {
            \includegraphics[height=.15\linewidth, width=.15\linewidth]{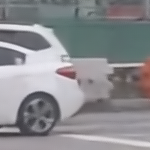}
        }
        \hfill
        \subfloat[\scriptsize R2R~\cite{R2R}]
        {
            \includegraphics[height=.15\linewidth, width=.15\linewidth]{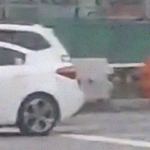}
        }
        
        \vspace{-2mm}
        \subfloat[\scriptsize Blurry Image]
        {
            \includegraphics[height=.15\linewidth, width=.15\linewidth]{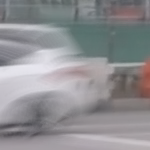}
        }
        \hfill
        \subfloat[\scriptsize Baseline$_\mathcal{B}$]
        {
            \includegraphics[height=.15\linewidth, width=.15\linewidth]{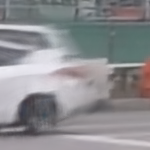}
        }
        \subfloat[\tiny Neighbor2Neighbor~\cite{Nei2Nei}]
        {
            \includegraphics[height=.15\linewidth, width=.15\linewidth]{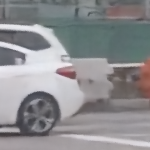}
        }
        \hfill
        \subfloat[\scriptsize Blind2Unblind~\cite{Blind2Unblind}]
        {
            \includegraphics[height=.15\linewidth, width=.15\linewidth]{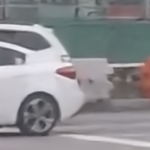}
        }
        \hfill
        \subfloat[\scriptsize SelfIR (Ours)]
        {
            \includegraphics[height=.15\linewidth, width=.15\linewidth]{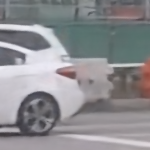}
        }
        \hfill
        \subfloat[\scriptsize Baseline$_\mathcal{R}$]
        {
            \includegraphics[height=.15\linewidth, width=.15\linewidth]{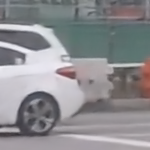}
        }
    \end{minipage}
    \vspace{-1mm}
    \caption{Visual comparison on Poisson noise. In terms of the visual result, SelfIR combines the advantages of supervised denoising Baseline$_\mathcal{N}$ and deblurring Baseline$_\mathcal{B}$.}
    \label{fig:poi1}
    \vspace{-3mm}
\end{figure}

\begin{table}[t]
  \small
  \renewcommand\arraystretch{1}
  \begin{center}
    \caption{Quantitative results on synthetic and real-world raw-RGB images.}
    \label{table:raw}
    \vspace{2mm}
    \scalebox{1}{
    \begin{tabular}{cccccc}
      \toprule
      & Method
      & \tabincell{c}{Sensor Noise~\cite{UPI} \\ \footnotesize{PSNR}\scriptsize{$\uparrow$} \small{/} \footnotesize{SSIM}\scriptsize{$\uparrow$} \small{/} \footnotesize{LPIPS}\scriptsize{$\downarrow$}}
      & \tabincell{c}{Real-World Images  \\ \footnotesize{NIQE}\scriptsize{$\downarrow$} \small{/} \footnotesize{NRQM}\scriptsize{$\uparrow$} \small{/} \footnotesize{PI}\scriptsize{$\downarrow$}} \\

      \midrule
      \multirow{2}{*}{\begin{tabular}[c]{@{}c@{}}Supervised\\Deblurring \end{tabular}}
      & Baseline$_\mathcal{B}$  & 28.14 / 0.8547 / 0.162   & 6.26 / 5.04 / 5.62 \\
      & DeepDeblur~\cite{DeepDeblur}  & 29.75 / 0.8881 / 0.115  & 6.76 / 4.78 / 6.00 \\

      \midrule
      \multirow{2}{*}{\begin{tabular}[c]{@{}c@{}}Supervised\\Denoising \end{tabular}}
      & Baseline$_\mathcal{N}$  & 34.52 / 0.9461 / 0.053  & 5.69 / 4.85 / 5.43 \\
      & DnCNN~\cite{DnCNN}  & 33.81 / 0.9325 / 0.076 & 6.05 / 5.10 / 5.48 \\

      \midrule
      Supervised IR & Baseline$_\mathcal{R}$       & 36.10 / 0.9574 / 0.035  & 5.54 / 5.14 / 5.18 \\

      \midrule
      \multirow{8}{*}{\begin{tabular}[c]{@{}c@{}}Self-Supervised\\Denoising \end{tabular}}
      & N2N~\cite{Noise2Noise}      & 34.67 / 0.9472 / 0.053 & 6.10 / 4.93 / 5.59 \\
      & N2V~\cite{N2V}              & 31.39 / 0.9227 / 0.076 & 5.82 / 5.52 / 5.17 \\
      & Laine19-mu~\cite{Laine19}   & 32.74 / 0.9304 / 0.073 & 5.87 / 5.67 / 5.10 \\
      & Laine19-pme~\cite{Laine19}  & 33.28 / 0.9119 / 0.095 & 7.26 / 6.03 / 5.62 \\
      & DBSN~\cite{DBSN}            & 33.59 / 0.9389 / 0.060 & 6.57 / 5.48 / 5.54 \\
      & R2R~\cite{R2R}              & 32.21 / 0.8807 / 0.117 & 5.63 / 5.63 / 4.99 \\
      & Neighbor2Neighbor~\cite{Nei2Nei}    & 32.82 / 0.9275 / 0.087  & 6.47 / 5.86 / 5.33 \\
      & Blind2Unblind~\cite{Blind2Unblind}  & 33.30 / 0.9380 / 0.061  & 5.28 / 5.22 / 5.04 \\ 

      \midrule
      \multirow{1}{*}{Ours}
      & SelfIR       & 34.51 / 0.9440 / 0.053 & 5.48 / 5.83 / 4.86 \\  
      \bottomrule
    \end{tabular}}
  \end{center}
  \vspace{-2mm}
\end{table}

\subsection{Experimental Results in Raw-RGB Space}
\cref{table:raw} shows the quantitative results of synthetic and real-world experiments on raw-RGB images.
For synthetic experiments with more complex and realistic sensor noise~\cite{UPI}, all models are retrained with our synthetic images.
Our SelfIR achieves a 1.69 dB PSNR gain in comparison with Neighbor2Neighbor~\cite{Nei2Nei}.
For evaluation on real-world images, we directly use the models of N2N~\cite{Noise2Noise}, Laine19-pme~\cite{Laine19}, R2R~\cite{R2R} and all supervised methods, which are pre-trained for sensor noise.
For SelfIR and other self-supervised methods, we fine-tune the pre-trained model on our real-world training set.
The PI~\cite{PI} metric of SelfIR is improved by 0.55 on real-world testing images through fine-tuning.
And the results on no-reference IQA show that our method is very competitive in comparison with the competing methods on real-world images.
The visual results are given in the supplemental material.

\begin{table}[t] 
  \small
  \vspace{1mm}
  \renewcommand\arraystretch{1}
  \begin{center}
    \caption{Results of deblurring with clear images and noisy images as the supervision.}
    \label{table:un_deblurring}
    \vspace{2mm}
    \scalebox{1}{
    \begin{tabular}{cccccc}
      \toprule
      \tabincell{c}{Supervision \\Information}
      & \tabincell{c}{Gaussian $\sigma \in [5/255,50/255]$ \\ \footnotesize{PSNR}\scriptsize{$\uparrow$} \small{/} \footnotesize{SSIM}\scriptsize{$\uparrow$} \small{/} \footnotesize{LPIPS}\scriptsize{$\downarrow$}}
      & \tabincell{c}{Poisson $\lambda \in [5,50]$  \\  \footnotesize{PSNR}\scriptsize{$\uparrow$} \small{/} \footnotesize{SSIM}\scriptsize{$\uparrow$} \small{/} \footnotesize{LPIPS}\scriptsize{$\downarrow$}} 
      & \tabincell{c}{Sensor Noise~\cite{UPI} \\  \footnotesize{PSNR}\scriptsize{$\uparrow$} \small{/} \footnotesize{SSIM}\scriptsize{$\uparrow$} \small{/} \footnotesize{LPIPS}\scriptsize{$\downarrow$}} \\
      \midrule
      Clear Images  & 28.24 / 0.8561 / 0.191 & 28.24 / 0.8561 / 0.191 & 28.14 / 0.8547 / 0.162  \\
      Noisy Images   & 28.29 / 0.8578 / 0.190  & 28.23 / 0.8563 / 0.191  & 28.16 / 0.8545 / 0.164 \\
      \bottomrule
    \end{tabular}}
  \end{center}
\end{table}

\begin{table}[t!] 
  \small
  \renewcommand\arraystretch{1}
  \begin{center}
    \caption{Ablation study of auxiliary loss on Gaussian noise.}
    \label{table:aux_loss}
    \vspace{2mm}
    \scalebox{1}{
    \begin{tabular}{cccccc}
      \toprule
      & \tabincell{c}{Neighbor2Neighbor~\cite{Nei2Nei} \\ \footnotesize{PSNR}\scriptsize{$\uparrow$} \small{/} \footnotesize{SSIM}\scriptsize{$\uparrow$} \small{/} \footnotesize{LPIPS}\scriptsize{$\downarrow$}}
      & \tabincell{c}{SelfIR \\ \footnotesize{PSNR}\scriptsize{$\uparrow$} \small{/} \footnotesize{SSIM}\scriptsize{$\uparrow$} \small{/} \footnotesize{LPIPS}\scriptsize{$\downarrow$}} \\
      \midrule
      w/o $\mathcal{L}_{aux}$ & 34.29 / 0.9271 / 0.085 & 35.65 / 0.9492 / 0.080 \\
      w/ $\mathcal{L}_{aux}$ & 34.45 / 0.9307 / 0.093 & 35.74 / 0.9499 / 0.076  \\
      \bottomrule
    \end{tabular}}
  \end{center}
\end{table}

\section{Ablation Study}

\subsection{Feasibility of Deblurring with Noisy Images}
In order to verify the feasibility of taking noisy images as the supervision of deblurring in \cref{sec:3.1}, we replace the clear supervision of Baseline$_\mathcal{B}$ with noisy images of different noise distributions, and the results are shown in \cref{table:un_deblurring}.
It can be seen that taking noisy or clear images as the supervision leads to similar performance on the deblurring task, which confirms that the noisy images can be an alternative of clear images to supervise the deblurring task.

\subsection{Effect of Auxiliary Loss}
In order to evaluate the effect of auxiliary loss $\mathcal{L}_{aux}$ in \cref{loss:aux}, we add $\mathcal{L}_{aux}$ to the loss terms of self-supervised denoising method Neighbor2Neighbor~\cite{Nei2Nei}.
As shown in \cref{table:aux_loss}, we obtain 0.16 dB PSNR gain against the baseline Neighbor2Neighbor~\cite{Nei2Nei}.
The result indicates that blurry images can provide some auxiliary supervision information and improve performance for self-supervised denoising.
When removing $\mathcal{L}_{aux}$ from our SelfIR, the PSNR dropped by 0.09 dB.
Since the long-exposure image is taken directly as input, the loss term may be less effective, but is still beneficial.

\begin{minipage}[t]{\linewidth}
    \small
    \centering
    \begin{minipage}[t]{0.45\linewidth}
        \centering
        \captionof{table}{\centering Ablation study  on different weights ($\lambda_{reg}$ values) of regularization loss.}
        \label{table:reg_loss_w}
        \begin{tabular}{cc}
            \toprule
            $\lambda_{reg}$   & \footnotesize{PSNR}\scriptsize{$\uparrow$} \small{/} \footnotesize{SSIM}\scriptsize{$\uparrow$} \small{/} \footnotesize{LPIPS}\scriptsize{$\downarrow$} \\
            \midrule
            0  & 35.20 / 0.9473 / 0.097 \\
            1  & 35.64 / 0.9492 / 0.082 \\
            2  & 35.74 / 0.9499 / 0.076 \\
            4  &  35.72 / 0.9496 / 0.075 \\
            8  &  35.73 / 0.9497 / 0.072  \\
           \bottomrule
        \end{tabular}
    \end{minipage}
    \hspace{3mm}
    \begin{minipage}[t]{0.45\linewidth}
        \centering
        \captionof{table}{\centering Ablation study on different weights ($\lambda_{aux}$ values) of auxiliary loss.}
        \label{table:aux_loss_w}
        \begin{tabular}{cc}
            \toprule
            $\lambda_{aux}$   & \footnotesize{PSNR}\scriptsize{$\uparrow$} \small{/} \footnotesize{SSIM}\scriptsize{$\uparrow$} \small{/} \footnotesize{LPIPS}\scriptsize{$\downarrow$} \\
            \midrule
            0  & 35.65 / 0.9492 / 0.080 \\
            1  & 35.73 / 0.9498 / 0.078 \\
            2  & 35.74 / 0.9499 / 0.076 \\
            4  &  35.73 / 0.9499 / 0.076 \\
            8  &  35.67 / 0.9496 / 0.077 \\
           \bottomrule
        \end{tabular}
    \end{minipage}
\end{minipage}
\vspace{3mm}

\subsection{Effect of Different Loss Weights}
We conduct ablation studies on different weighting hyper-parameters (\ie, $\lambda_{reg}$ and $\lambda_{reg}$) for balancing regularization and auxiliary loss terms.
%
%
The experiments are conducted on Gaussian noise in sRGB space.
When varying one hyper-parameter, the other one is set to 2 by default.
As shown in Tabs.~\ref{table:reg_loss_w} and ~\ref{table:aux_loss_w}, it can be seen that the sensitivity to $\lambda_{reg}$ and $\lambda_{reg}$ of SelfIR is acceptable.

\section{Conclusion}
The complementarity between long-exposure blurry and short-exposure noisy images not only improves the performance of image restoration, but also makes it possible to learn a restoration model in a self-supervised manner.
Jointly leveraging the long- and short-exposure images, we present a self-supervised image restoration method named SelfIR.
On the one hand, we take the short-exposure images as the supervision information for deblurring.
On the other hand, we utilize the sharp areas in the long-exposure images as auxiliary supervision information in aid of self-supervised denoising.
SelfIR combines these two aspects by a collaborative learning scheme, and makes the deblurring and denoising tasks benefit from each other. 
Experiments on synthetic and real-world images show the effectiveness and practicality of our proposed method. 

\section{Limitation, Impact, Etc}\label{sec:limit}
This work is still limited in assessing the results on real-world image pairs.
Although no-reference IQA metrics are adopted, these may be too unstable to report the performance consistent with humans accurately.
We will establish a real-world blurry-noisy pair dataset including high-quality GTs to solve this problem in the future work.
As for societal influence, this work is promising to be applied to terminal devices (\eg., smartphones) for obtaining better images under low-light environments.
It has no foreseeable negative impact.
Besides, the images used in this work are from natural or human scenes.
The GoPro dataset used for synthetic experiments is public under CC BY 4.0 license.
There is no personally identifiable information or offensive content in the experimental data.

\section*{Acknowledgement}
This work was supported by the Major Key Project of Peng Cheng Laboratory (PCL2021A12) and the National Natural Science Foundation of China (NSFC) under Grants No.s U19A2073 and 61871381.

\clearpage


\appendix

\section*{\centering{\Large Appendix\\[30pt]}}

\section{Content}
  The content of the supplementary material involves:
  \begin{itemize}
    \item  Formulations of synthetic noise in Sec.~\ref{sec:noise}.  %
    \item  Detailed collection process of real-world dataset in Sec.~\ref{sec:dataset}.  %
    \item Detailed architecture of restoration network  in Sec.~\ref{sec:network}. %
    \item More  qualitative results  in Sec.~\ref{sec:visual}. %
  \end{itemize}

\section{Formulations of Synthetic Noise} \label{sec:noise}
\noindent\textbf{Gaussian Noise.}
Sensor noise mainly includes shot and read noise, which are caused by photon arrival statistics biases and imprecise readout circuits, respectively.
Gaussian noise is a good approximation of read noise and is widely used as the assumption for denoising.
In the experiments, the synthetic Gaussian noise $\mathbf{N}_\mathcal{N}$ for the short-exposure image can be expressed as,
\begin{equation}
 \mathbf{N}_\mathcal{N} \sim \mathcal{N}  (\mathbf{0}, \sigma^2\mathbf{E}) 
\end{equation}
where $ \mathcal{N}(\cdot,\cdot) $ denotes Gaussian  distribution, $ \mathbf {E} $ is an identity matrix with ones on the main diagonal and zeros elsewhere.
Following previous works~\cite{Nei2Nei,Blind2Unblind}, the standard deviation $\sigma$ is chosen uniformly at random from $[5/255, 50/255]$ when color intensity values of images are normalized to $[0, 1]$.

\noindent\textbf{Poisson Noise.}
During imaging, the sensor receives photons and performs the photoelectric conversion.
The quantum nature of light leads to uncertainty in the number of electrons, which introduces shot noise in the image.
The shot noise follows a Poisson distribution $\mathcal {P}$ over the number of electrons, thus we also experiment using the noisy image $\mathbf{I}_\mathcal{N}$ with Poisson noise, which can be expressed as,
\begin{equation}
\mathbf{I}_\mathcal{N} \sim \mathcal{P} (\lambda \cdot \mathbf{I}) / \lambda,
\end{equation}
where $\mathbf{I}$ denotes a clean image, $\lambda\!\in\![5, 50]$ according to ~\cite{Nei2Nei,Blind2Unblind}.

\noindent\textbf{Sensor Noise.} 
Recently, Brook~\etal~\cite{UPI} propose to approximate shot and read noise as a single heteroscedastic Gaussian noise.
The noise is more complex but more realistic than homoscedastic Gaussian or Poisson noise, which can be written as,
\begin{equation}
\mathbf{N}_\mathcal{N} \sim \mathcal{N}(\mathbf{0}, \lambda_{read} + \lambda_{shot} \mathbf{I}).
\end{equation}
where $\lambda_{read}$ and $\lambda_{shot}$ are determined by modeling the joint distribution of different shot/read noise parameter pairs in real-world raw images and sampling from that distribution.
For SIDD~\cite{SIDD} dataset, the sampling shot/read noise factors can be expressed as,
\begin{equation}
\begin{split}
& \log(\lambda_{shot}) \sim {\mathcal{U}}(\log(0.00068674), \log(0.02194856)), \\
& \log(\lambda_{read}) \mid \log(\lambda_{shot}) \sim {\mathcal {N}}(1.85 \lambda_{shot} + 0.3, 0.2^2).
\end{split}
\end{equation}
where $\mathcal{U}(a,b)$ denotes a uniform distribution in the range $[a,b]$.

\section{Detailed Collection Process of Real-World Dataset} \label{sec:dataset}
For obtaining real-world blurry-noisy image pairs, we utilize two Huawei P40 smartphones to capture a short-exposure noisy image and a long-exposure blurry image of the same scene, respectively.
In the pair, the product of ISO and exposure time of two images remains the same, \ie, the noisy image is captured with a high ISO and short exposure time, while the blurry one is captured with a low ISO and long exposure time.
In order to make the captured image pair align as much as possible, the two smartphones take images side by side at the same time.
Nonetheless, spatial misalignment still exists between the image pair, as the two camera lenses are not in the same position.

\begin{figure}[t]
    \begin{minipage}[t]{\linewidth}
        \centering
        \tiny
        \subfloat[]
        {
            \includegraphics[width=.48\linewidth]{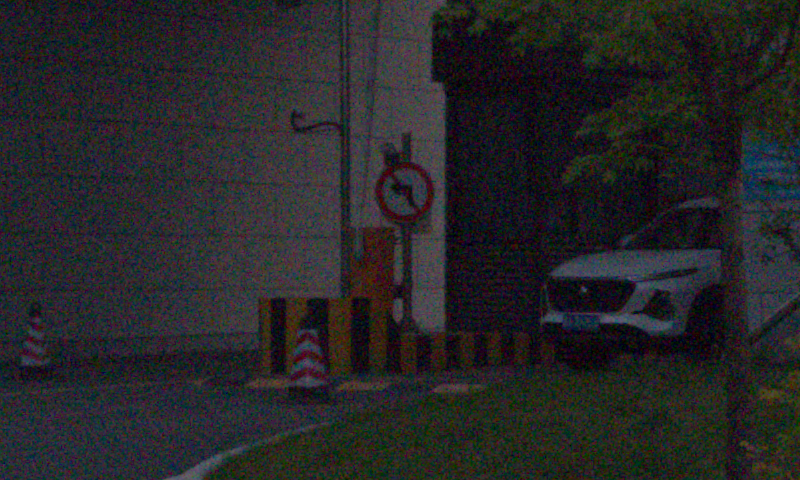}
        }
        \hfill
        \subfloat[]
        {
            \includegraphics[width=.48\linewidth]{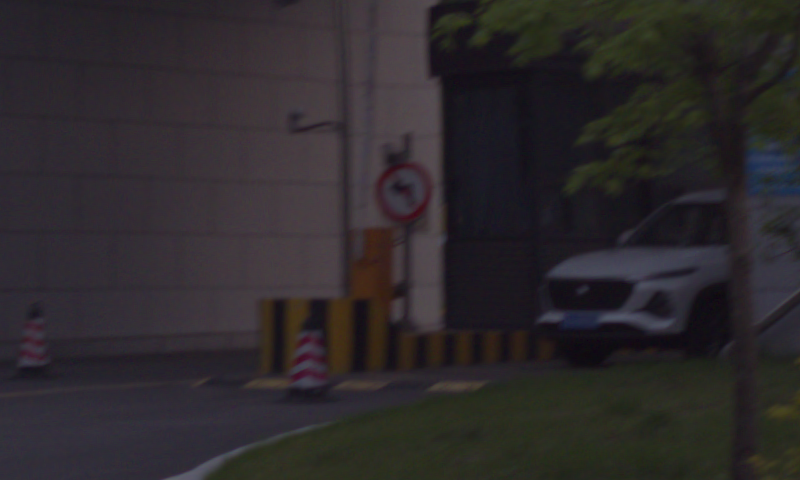}
        }
        
        \subfloat[Short-Exposure Noisy Image]
        {
            \includegraphics[width=.48\linewidth]{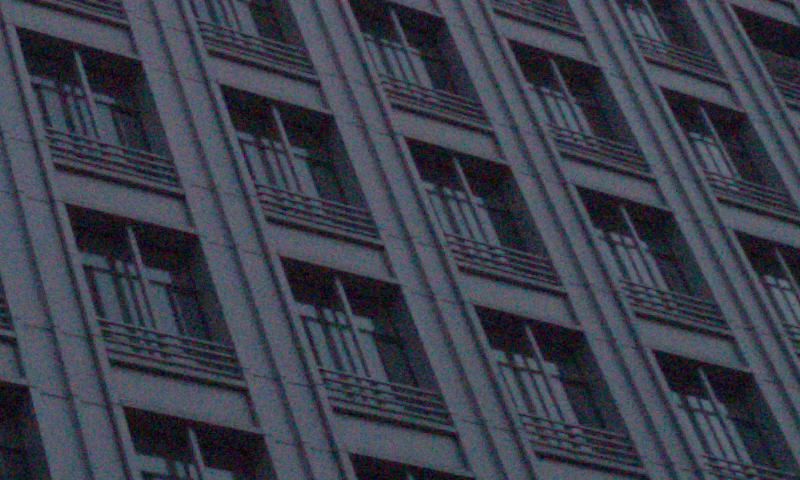}
        }
        \hfill
        \subfloat[Long-Exposure Blurry Image]
        {
            \includegraphics[width=.48\linewidth]{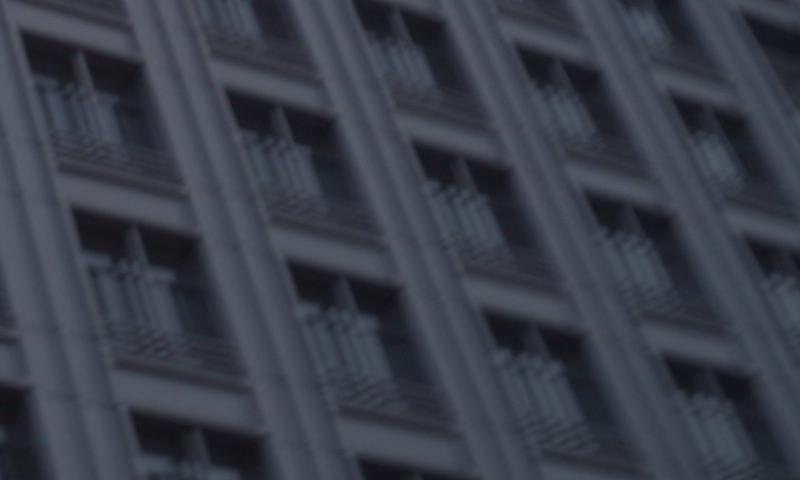}
        }
    \end{minipage}
    \caption{Example image patches of our real-world blurry-noisy pairs.}
    \label{fig:real_data}
\end{figure}

Therefore, we further exploit the optical flow network PWC-Net~\cite{PWC-Net} to align the image pair.
First, the raw-RGB pair is converted to sRGB space through camera image signal processing (ISP) pipeline.
Then we calculate the optical flow between the sRGB image pair and use the optical flow to warp raw-RGB blurry image.
Finally, the approximately aligned blurry-noisy raw-RGB pair can be obtained.
Fig.~\ref{fig:real_data} shows two examples of our real-world blurry-noisy image pairs.

\section{Detailed Architecture of Restoration Network} \label{sec:network}
\begin{figure}[t]
	\centering
	\begin{overpic}[width=0.99\linewidth]{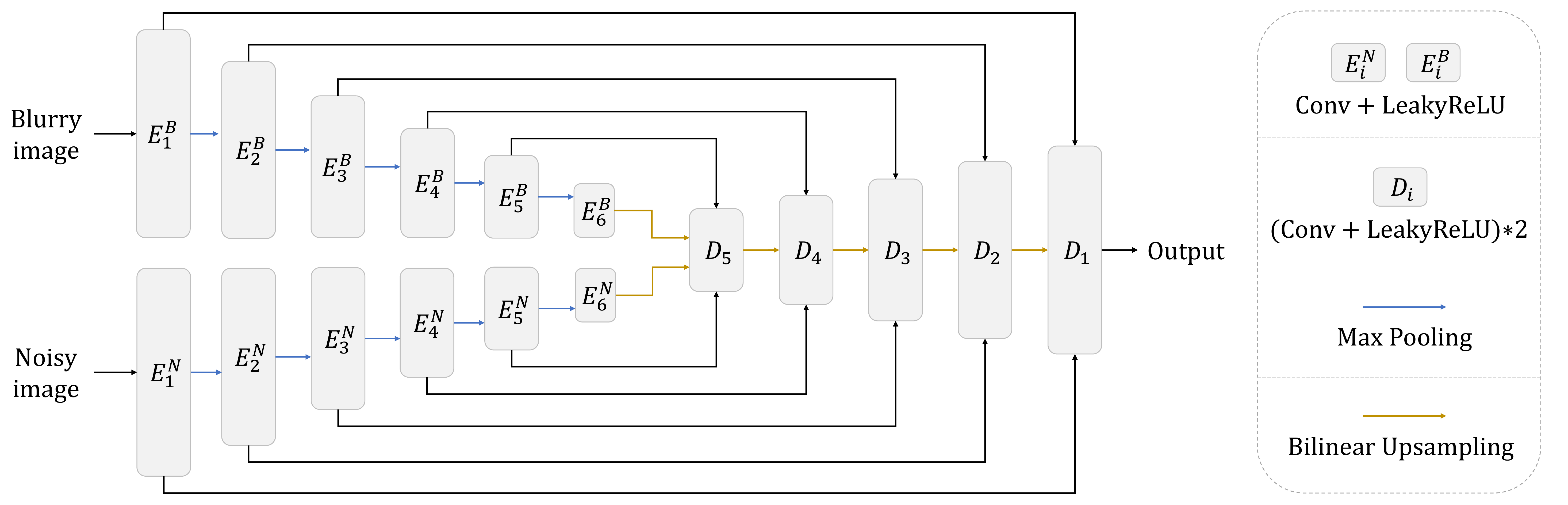}
	\end{overpic}
	\caption{Architecture of restoration network. $E_i^B$  and $E_i^N$ denote the $i$-th level of encoder for noisy image and blurry image, respectively. $D_i$ denotes the $i$-th level of decoder.}
	\label{fig:network}
\end{figure}

Fig.~\ref{fig:network} shows the detailed architecture of restoration network.
Two encoders are respectively deployed to the blurry and noisy input images for better domain-specific feature extraction, and then blurry and noisy features are fused in the decoder part.
The architectures of encoder and decoder are the same with these of previous self-supervised denoising methods~\cite{Noise2Noise,Laine19,Nei2Nei}.

\section{More Qualitative Results} \label{sec:visual}
\begin{figure}[t!]
    \vspace{-2mm}
    \begin{minipage}[t]{\linewidth}
        \centering
        \tiny
        \subfloat[\scriptsize Noisy Image]
        {
            \includegraphics[height=.15\linewidth, width=.15\linewidth]{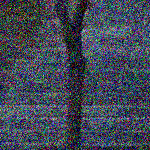}
        }
        \hfill
        \subfloat[\scriptsize Baseline$_\mathcal{N}$]
        {
            \includegraphics[height=.15\linewidth, width=.15\linewidth]{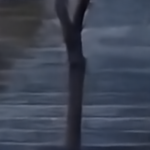}
        }
        \hfill
        \subfloat[\scriptsize N2V~\cite{N2V}]
        {
            \includegraphics[height=.15\linewidth, width=.15\linewidth]{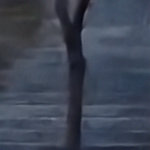}
        }
        \hfill
        \subfloat[\scriptsize Laine19-pme~\cite{Laine19}]
        {
            \includegraphics[height=.15\linewidth, width=.15\linewidth]{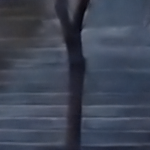}
        }
        \hfill
        \subfloat[\scriptsize DBSN~\cite{DBSN}]
        {
            \includegraphics[height=.15\linewidth, width=.15\linewidth]{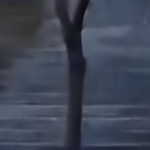}
        }
        \hfill
        \subfloat[\scriptsize R2R~\cite{R2R}]
        {
            \includegraphics[height=.15\linewidth, width=.15\linewidth]{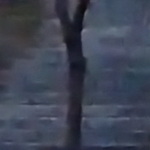}
        }
        \vspace{-2mm}

        \subfloat[\scriptsize Blurry Image]
        {
            \includegraphics[height=.15\linewidth, width=.15\linewidth]{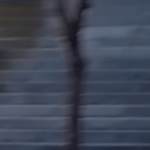}
        }
        \hfill
        \subfloat[\scriptsize Baseline$_\mathcal{B}$]
        {
            \includegraphics[height=.15\linewidth, width=.15\linewidth]{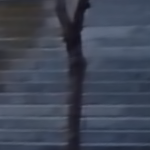}
        }
        \hfill        
        \subfloat[\tiny Neighbor2Neighbor~\cite{Nei2Nei}]
        {
            \includegraphics[height=.15\linewidth, width=.15\linewidth]{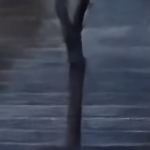}
        }
        \hfill
        \subfloat[\scriptsize Blind2Unblind~\cite{Blind2Unblind}]
        {
            \includegraphics[height=.15\linewidth, width=.15\linewidth]{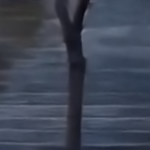}
        }
        \hfill
        \subfloat[\scriptsize SelfIR (Ours)]
        {
            \includegraphics[height=.15\linewidth, width=.15\linewidth]{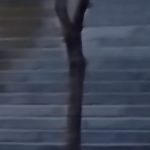}
        }
        \hfill
        \subfloat[\scriptsize GT]
        {
            \includegraphics[height=.15\linewidth, width=.15\linewidth]{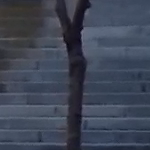}
        }
    \end{minipage}
    \vspace{-1mm}
    \caption{Visual comparison on Gaussian noise.  The  edges of tree and steps in our result are sharper.}
    \label{fig:gau2}
\end{figure}

\begin{figure}[t]
    \vspace{-2mm}
    \begin{minipage}[t]{\linewidth}
        \centering
        \tiny    
        \subfloat[\scriptsize Noisy Image]
        {
            \includegraphics[height=.15\linewidth, width=.15\linewidth]{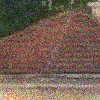}
        }
        \hfill
        \subfloat[\scriptsize Baseline$_\mathcal{N}$]
        {
            \includegraphics[height=.15\linewidth, width=.15\linewidth]{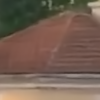}
        }
        \hfill
        \subfloat[\scriptsize N2V~\cite{N2V}]
        {
            \includegraphics[height=.15\linewidth, width=.15\linewidth]{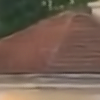}
        }
        \hfill
        \subfloat[\scriptsize Laine19-pme~\cite{Laine19}]
        {
            \includegraphics[height=.15\linewidth, width=.15\linewidth]{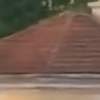}
        }
        \hfill
        \subfloat[\scriptsize DBSN~\cite{DBSN}]
        {
            \includegraphics[height=.15\linewidth, width=.15\linewidth]{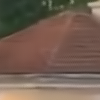}
        }
        \hfill
        \subfloat[\scriptsize R2R~\cite{R2R}]
        {
            \includegraphics[height=.15\linewidth, width=.15\linewidth]{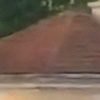}
        }
        
        \vspace{-2mm}
        \subfloat[\scriptsize Blurry Image]
        {
            \includegraphics[height=.15\linewidth, width=.15\linewidth]{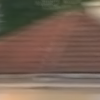}
        }
        \hfill
        \subfloat[\scriptsize Baseline$_\mathcal{B}$]
        {
            \includegraphics[height=.15\linewidth, width=.15\linewidth]{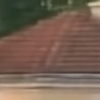}
        }
        \hfill
        \subfloat[\tiny Neighbor2Neighbor~\cite{Nei2Nei}]
        {
            \includegraphics[height=.15\linewidth, width=.15\linewidth]{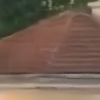}
        }
        \hfill
        \subfloat[\scriptsize Blind2Unblind~\cite{Blind2Unblind}]
        {
            \includegraphics[height=.15\linewidth, width=.15\linewidth]{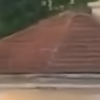}
        }
        \hfill
        \subfloat[\scriptsize SelfIR (Ours)]
        {
            \includegraphics[height=.15\linewidth, width=.15\linewidth]{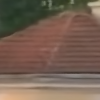}
        }
        \hfill
        \subfloat[\scriptsize GT]
        {
            \includegraphics[height=.15\linewidth, width=.15\linewidth]{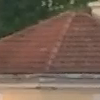}
        }
    \end{minipage}
    \vspace{0mm}
    \caption{Visual comparison on Poisson noise. The textures in our result are clearer and more regular.}
    \label{fig:poi2}
    \vspace{1mm}
\end{figure}

\begin{figure}[t]
    \vspace{-2mm}
    \begin{minipage}[t]{\linewidth}
        \centering
        \tiny
        \subfloat[\scriptsize Noisy Image]
        {
            \includegraphics[height=.15\linewidth, width=.15\linewidth]{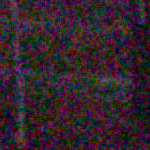}
        }
        \hfill
        \subfloat[\scriptsize Baseline$_\mathcal{N}$]
        {
            \includegraphics[height=.15\linewidth, width=.15\linewidth]{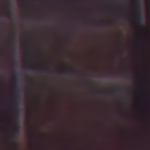}
        }
        \hfill
        \subfloat[\scriptsize N2V~\cite{N2V}]
        {
            \includegraphics[height=.15\linewidth, width=.15\linewidth]{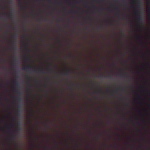}
        }
        \hfill
        \subfloat[\scriptsize Laine19-pme~\cite{Laine19}]
        {
            \includegraphics[height=.15\linewidth, width=.15\linewidth]{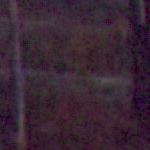}
        }
        \hfill
        \subfloat[\scriptsize DBSN~\cite{DBSN}]
        {
            \includegraphics[height=.15\linewidth, width=.15\linewidth]{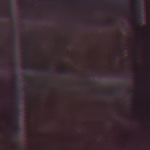}
        }
        \hfill
        \subfloat[\scriptsize R2R~\cite{R2R}]
        {
            \includegraphics[height=.15\linewidth, width=.15\linewidth]{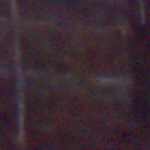}
        }
        
        \vspace{-2mm}
        \subfloat[\scriptsize Blurry Image]
        {
            \includegraphics[height=.15\linewidth, width=.15\linewidth]{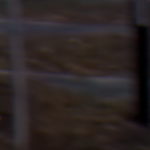}
        }
        \hfill
        \subfloat[\scriptsize Baseline$_\mathcal{B}$]
        {
            \includegraphics[height=.15\linewidth, width=.15\linewidth]{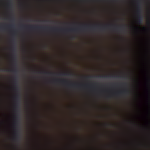}
        }
        \hfill
        \subfloat[\tiny Neighbor2Neighbor~\cite{Nei2Nei}]
        {
            \includegraphics[height=.15\linewidth, width=.15\linewidth]{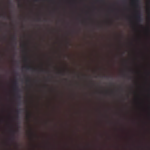}
        }
        \hfill
        \subfloat[\scriptsize Blind2Unblind~\cite{Blind2Unblind}]
        {
            \includegraphics[height=.15\linewidth, width=.15\linewidth]{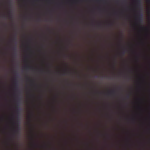}
        }
        \hfill
        \subfloat[\scriptsize SelfIR (Ours)]
        {
            \includegraphics[height=.15\linewidth, width=.15\linewidth]{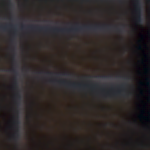}
        }
        \hfill
        \subfloat[\scriptsize GT]
        {
            \includegraphics[height=.15\linewidth, width=.15\linewidth]{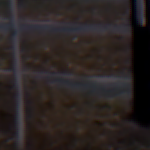}
        }
    \end{minipage}
    \vspace{0mm}
    \caption{Visual comparison on sensor noise. Our result is more photo-realistic and faithfully recovers the image color.}
    \label{fig:sensor1}
    \vspace{1mm}
\end{figure}

\begin{figure}[t]
    \vspace{-2mm}
    \begin{minipage}[t]{\linewidth}
        \centering
        \tiny
        \subfloat[\scriptsize Noisy Image]
        {
            \includegraphics[height=.15\linewidth, width=.15\linewidth]{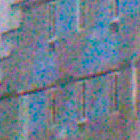}
        }
        \hfill
        \subfloat[\scriptsize Baseline$_\mathcal{N}$]
        {
            \includegraphics[height=.15\linewidth, width=.15\linewidth]{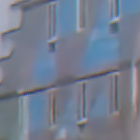}
        }
        \hfill
        \subfloat[\scriptsize N2V~\cite{N2V}]
        {
            \includegraphics[height=.15\linewidth, width=.15\linewidth]{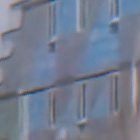}
        }
        \hfill
        \subfloat[\scriptsize Laine19-pme~\cite{Laine19}]
        {
            \includegraphics[height=.15\linewidth, width=.15\linewidth]{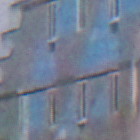}
        }
        \hfill
        \subfloat[\scriptsize DBSN~\cite{DBSN}]
        {
            \includegraphics[height=.15\linewidth, width=.15\linewidth]{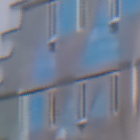}
        }
        \hfill
        \subfloat[\scriptsize R2R~\cite{R2R}]
        {
            \includegraphics[height=.15\linewidth, width=.15\linewidth]{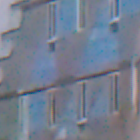}
        }
        
        \vspace{-2mm}
        \subfloat[\scriptsize Blurry Image]
        {
            \includegraphics[height=.15\linewidth, width=.15\linewidth]{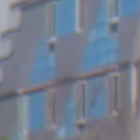}
        }
        \hfill
        \subfloat[\scriptsize Baseline$_\mathcal{B}$]
        {
            \includegraphics[height=.15\linewidth, width=.15\linewidth]{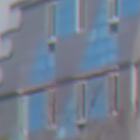}
        }
        \hfill
        \subfloat[\tiny Neighbor2Neighbor~\cite{Nei2Nei}]
        {
            \includegraphics[height=.15\linewidth, width=.15\linewidth]{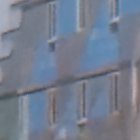}
        }
        \hfill
        \subfloat[\scriptsize Blind2Unblind~\cite{Blind2Unblind}]
        {
            \includegraphics[height=.15\linewidth, width=.15\linewidth]{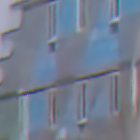}
        }
        \hfill
        \subfloat[\scriptsize SelfIR (Ours)]
        {
            \includegraphics[height=.15\linewidth, width=.15\linewidth]{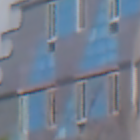}
        }
        \hfill
        \subfloat[\scriptsize GT]
        {
            \includegraphics[height=.15\linewidth, width=.15\linewidth]{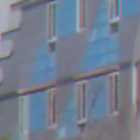}
        }
    \end{minipage}
    \vspace{0mm}
    \caption{Visual comparison on sensor noise. Our SelfIR recovers much more details.}
    \label{fig:sensor2}
    \vspace{1mm}
\end{figure}

\begin{figure}[t]
    \vspace{-2mm}
    \begin{minipage}[t]{\linewidth}
        \centering
        \tiny
        \subfloat[\scriptsize Noisy Image]
        {
            \includegraphics[height=.15\linewidth, width=.15\linewidth]{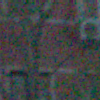}
        }
        \hfill
        \subfloat[\scriptsize Baseline$_\mathcal{N}$]
        {
            \includegraphics[height=.15\linewidth, width=.15\linewidth]{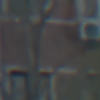}
        }
        \hfill
        \subfloat[\scriptsize N2V~\cite{N2V}]
        {
            \includegraphics[height=.15\linewidth, width=.15\linewidth]{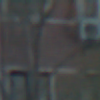}
        }
        \hfill
        \subfloat[\scriptsize Laine19-pme~\cite{Laine19}]
        {
            \includegraphics[height=.15\linewidth, width=.15\linewidth]{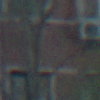}
        }
        \hfill
        \subfloat[\scriptsize DBSN~\cite{DBSN}]
        {
            \includegraphics[height=.15\linewidth, width=.15\linewidth]{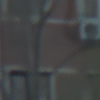}
        }
        \hfill
        \subfloat[\scriptsize R2R~\cite{R2R}]
        {
            \includegraphics[height=.15\linewidth, width=.15\linewidth]{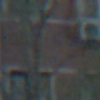}
        }
        
        \vspace{-2mm}
        \subfloat[\scriptsize Blurry Image]
        {
            \includegraphics[height=.15\linewidth, width=.15\linewidth]{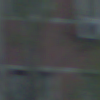}
        }
        \hfill
        \subfloat[\scriptsize Baseline$_\mathcal{B}$]
        {
            \includegraphics[height=.15\linewidth, width=.15\linewidth]{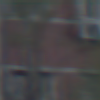}
        }
        \hfill
        \subfloat[\tiny Neighbor2Neighbor~\cite{Nei2Nei}]
        {
            \includegraphics[height=.15\linewidth, width=.15\linewidth]{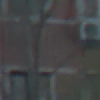}
        }
        \hfill
        \subfloat[\scriptsize Blind2Unblind~\cite{Blind2Unblind}]
        {
            \includegraphics[height=.15\linewidth, width=.15\linewidth]{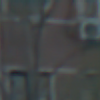}
        }
        \hfill
        \subfloat[\scriptsize SelfIR (Ours)]
        {
            \includegraphics[height=.15\linewidth, width=.15\linewidth]{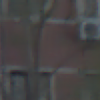}
        }
        \hfill
        \subfloat[\scriptsize  Baseline$_\mathcal{R}$]
        {
            \includegraphics[height=.15\linewidth, width=.15\linewidth]{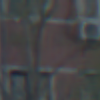}
        }
    \end{minipage}
    \vspace{0mm}
    \caption{Visual comparison on real-world images. Our result is clearer and more photo-realistic.}
    \label{fig:real1}
    \vspace{1mm}
\end{figure}

\begin{figure}[t]
    \vspace{-2mm}
    \begin{minipage}[t]{\linewidth}
        \centering
        \tiny
        \subfloat[\scriptsize Noisy Image]
        {
            \includegraphics[height=.15\linewidth, width=.15\linewidth]{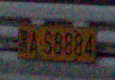}
        }
        \hfill
        \subfloat[\scriptsize Baseline$_\mathcal{N}$]
        {
            \includegraphics[height=.15\linewidth, width=.15\linewidth]{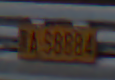}
        }
        \hfill
        \subfloat[\scriptsize N2V~\cite{N2V}]
        {
            \includegraphics[height=.15\linewidth, width=.15\linewidth]{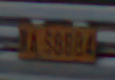}
        }
        \hfill
        \subfloat[\scriptsize Laine19-pme~\cite{Laine19}]
        {
            \includegraphics[height=.15\linewidth, width=.15\linewidth]{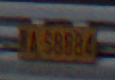}
        }
        \hfill
        \subfloat[\scriptsize DBSN~\cite{DBSN}]
        {
            \includegraphics[height=.15\linewidth, width=.15\linewidth]{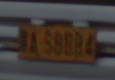}
        }
        \hfill
        \subfloat[\scriptsize R2R~\cite{R2R}]
        {
            \includegraphics[height=.15\linewidth, width=.15\linewidth]{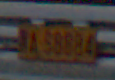}
        }
        
        \vspace{-2mm}
        \subfloat[\scriptsize Blurry Image]
        {
            \includegraphics[height=.15\linewidth, width=.15\linewidth]{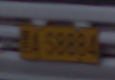}
        }
        \hfill
        \subfloat[\scriptsize Baseline$_\mathcal{B}$]
        {
            \includegraphics[height=.15\linewidth, width=.15\linewidth]{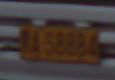}
        }
        \hfill
        \subfloat[\tiny Neighbor2Neighbor~\cite{Nei2Nei}]
        {
            \includegraphics[height=.15\linewidth, width=.15\linewidth]{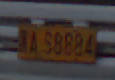}
        }
        \hfill
        \subfloat[\scriptsize Blind2Unblind~\cite{Blind2Unblind}]
        {
            \includegraphics[height=.15\linewidth, width=.15\linewidth]{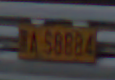}
        }
        \hfill
        \subfloat[\scriptsize SelfIR (Ours)]
        {
            \includegraphics[height=.15\linewidth, width=.15\linewidth]{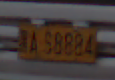}
        }
        \hfill
        \subfloat[\scriptsize Baseline$_\mathcal{R}$]
        {
            \includegraphics[height=.15\linewidth, width=.15\linewidth]{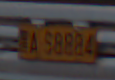}
        }
    \end{minipage}
    \vspace{0mm}
    \caption{Visual comparison on real-world images. The license plate in our result is clearer and sharper.}
    \label{fig:real2}
    \vspace{1mm}
\end{figure}

\cref{fig:gau1,fig:gau2} show more qualitative results  on Gaussian noise.
\cref{fig:poi1,fig:poi2} show more qualitative results  on Poisson noise.
The visual comparison on sensor noise~\cite{UPI} can be seen in \cref{fig:sensor1,fig:sensor2}.
The visual comparison on the real-world images can be seen in \cref{fig:real1,fig:real2}.
The results show our SelfIR performs favorably against the state-of-the-art self-supervised denoising methods, as well as the baseline supervised deblurring and denoising methods.

 \begin{figure}[t]
    \vspace{-2mm}
    \begin{minipage}[t]{\linewidth}
        \centering
        \tiny
        \subfloat[\scriptsize Noisy Image]
        {
            \includegraphics[height=.15\linewidth, width=.15\linewidth]{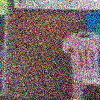}
        }
        \hfill
        \subfloat[\scriptsize Baseline$_\mathcal{N}$]
        {
            \includegraphics[height=.15\linewidth, width=.15\linewidth]{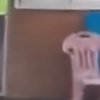}
        }
        \hfill
        \subfloat[\scriptsize N2V~\cite{N2V}]
        {
            \includegraphics[height=.15\linewidth, width=.15\linewidth]{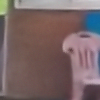}
        }
        \hfill
        \subfloat[\scriptsize Laine19-pme~\cite{Laine19}]
        {
            \includegraphics[height=.15\linewidth, width=.15\linewidth]{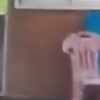}
        }
        \hfill
        \subfloat[\scriptsize DBSN~\cite{DBSN}]
        {
            \includegraphics[height=.15\linewidth, width=.15\linewidth]{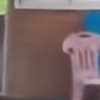}
        }
        \hfill
        \subfloat[\scriptsize R2R~\cite{R2R}]
        {
            \includegraphics[height=.15\linewidth, width=.15\linewidth]{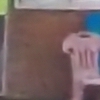}
        }
        \vspace{-2mm}

        \subfloat[\scriptsize Blurry Image]
        {
            \includegraphics[height=.15\linewidth, width=.15\linewidth]{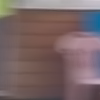}
        }
        \hfill
        \subfloat[\scriptsize Baseline$_\mathcal{B}$]
        {
            \includegraphics[height=.15\linewidth, width=.15\linewidth]{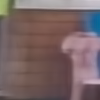}
        }
        \hfill        
        \subfloat[\tiny Neighbor2Neighbor~\cite{Nei2Nei}]
        {
            \includegraphics[height=.15\linewidth, width=.15\linewidth]{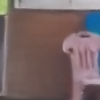}
        }
        \hfill
        \subfloat[\scriptsize Blind2Unblind~\cite{Blind2Unblind}]
        {
            \includegraphics[height=.15\linewidth, width=.15\linewidth]{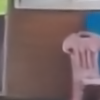}
        }
        \hfill
        \subfloat[\scriptsize SelfIR (Ours)]
        {
            \includegraphics[height=.15\linewidth, width=.15\linewidth]{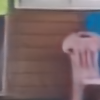}
        }
        \hfill
        \subfloat[\scriptsize GT]
        {
            \includegraphics[height=.15\linewidth, width=.15\linewidth]{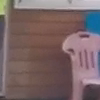}
        }
    \end{minipage}
    \vspace{-1mm}
    \caption{Visual comparison on Gaussian noise. The lines and chair in our result are clearer.}
    \label{fig:gau1}
\end{figure}

\clearpage

 \begin{figure}[t]
    \vspace{-2mm}
    \begin{minipage}[t]{\linewidth}
        \centering
        \tiny
        \subfloat[\scriptsize Noisy Image]
        {
            \includegraphics[height=.15\linewidth, width=.15\linewidth]{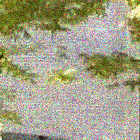}
        }
        \hfill
        \subfloat[\scriptsize Baseline$_\mathcal{N}$]
        {
            \includegraphics[height=.15\linewidth, width=.15\linewidth]{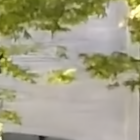}
        }
        \hfill
        \subfloat[\scriptsize N2V~\cite{N2V}]
        {
            \includegraphics[height=.15\linewidth, width=.15\linewidth]{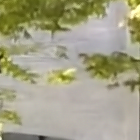}
        }
        \hfill
        \subfloat[\scriptsize Laine19-pme~\cite{Laine19}]
        {
            \includegraphics[height=.15\linewidth, width=.15\linewidth]{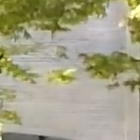}
        }
        \hfill
        \subfloat[\scriptsize DBSN~\cite{DBSN}]
        {
            \includegraphics[height=.15\linewidth, width=.15\linewidth]{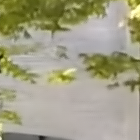}
        }
        \hfill
        \subfloat[\scriptsize R2R~\cite{R2R}]
        {
            \includegraphics[height=.15\linewidth, width=.15\linewidth]{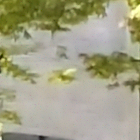}
        }
        
        \vspace{-2mm}
        \subfloat[\scriptsize Blurry Image]
        {
            \includegraphics[height=.15\linewidth, width=.15\linewidth]{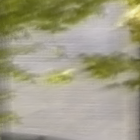}
        }
        \hfill
        \subfloat[\scriptsize Baseline$_\mathcal{B}$]
        {
            \includegraphics[height=.15\linewidth, width=.15\linewidth]{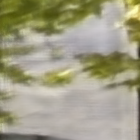}
        }
        \hfill
        \subfloat[\tiny Neighbor2Neighbor~\cite{Nei2Nei}]
        {
            \includegraphics[height=.15\linewidth, width=.15\linewidth]{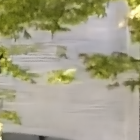}
        }
        \hfill
        \subfloat[\scriptsize Blind2Unblind~\cite{Blind2Unblind}]
        {
            \includegraphics[height=.15\linewidth, width=.15\linewidth]{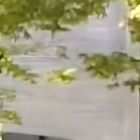}
        }
        \hfill
        \subfloat[\scriptsize SelfIR (Ours)]
        {
            \includegraphics[height=.15\linewidth, width=.15\linewidth]{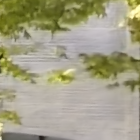}
        }
        \hfill
        \subfloat[\scriptsize GT]
        {
            \includegraphics[height=.15\linewidth, width=.15\linewidth]{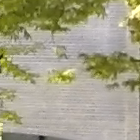}
        }
    \end{minipage}
    \vspace{0mm}
    \caption{Visual comparison on Poisson noise. Our SelfIR recovers more fine-scale textures.}
    \label{fig:poi1}
    \vspace{1mm}
\end{figure}        

\clearpage

{\small
\bibliographystyle{ieee_fullname}
\bibliography{egbib}

\begin{thebibliography}{10}\itemsep=-1pt

\bibitem{SIDD}
Abdelrahman Abdelhamed, Stephen Lin, and Michael~S Brown.
\newblock A high-quality denoising dataset for smartphone cameras.
\newblock In {\em CVPR}, 2018.

\bibitem{aittala2018burst}
Miika Aittala and Fr{\'e}do Durand.
\newblock Burst image deblurring using permutation invariant convolutional
  neural networks.
\newblock In {\em ECCV}, 2018.

\bibitem{PI}
Yochai Blau, Roey Mechrez, Radu Timofte, Tomer Michaeli, and Lihi Zelnik-Manor.
\newblock The 2018 pirm challenge on perceptual image super-resolution.
\newblock In {\em ECCV Workshops}, 2018.

\bibitem{UPI}
Tim Brooks, Ben Mildenhall, Tianfan Xue, Jiawen Chen, Dillon Sharlet, and
  Jonathan~T Barron.
\newblock Unprocessing images for learned raw denoising.
\newblock In {\em CVPR}, 2019.

\bibitem{Low-light-TMM}
Meng Chang, Huajun Feng, Zhihai Xu, and Qi Li.
\newblock Low-light image restoration with short-and long-exposure raw pairs.
\newblock {\em IEEE TMM}, 2021.

\bibitem{MIMO-UNet}
Sung-Jin Cho, Seo-Won Ji, Jun-Pyo Hong, Seung-Won Jung, and Sung-Jea Ko.
\newblock Rethinking coarse-to-fine approach in single image deblurring.
\newblock In {\em ICCV}, 2021.

\bibitem{delbracio2021mobile}
Mauricio Delbracio, Damien Kelly, Michael~S Brown, and Peyman Milanfar.
\newblock Mobile computational photography: A tour.
\newblock {\em Annual Review of Vision Science}, 2021.

\bibitem{fergus2006removing}
Rob Fergus, Barun Singh, Aaron Hertzmann, Sam~T Roweis, and William~T Freeman.
\newblock Removing camera shake from a single photograph.
\newblock In {\em ACM SIGGRAPH}, 2006.

\bibitem{CBDNet}
Shi Guo, Zifei Yan, Kai Zhang, Wangmeng Zuo, and Lei Zhang.
\newblock Toward convolutional blind denoising of real photographs.
\newblock In {\em CVPR}, 2019.

\bibitem{HDRplus}
Samuel~W Hasinoff, Dillon Sharlet, Ryan Geiss, Andrew Adams, Jonathan~T Barron,
  Florian Kainz, Jiawen Chen, and Marc Levoy.
\newblock Burst photography for high dynamic range and low-light imaging on
  mobile cameras.
\newblock {\em ACM ToG}, 2016.

\bibitem{Nei2Nei}
Tao Huang, Songjiang Li, Xu Jia, Huchuan Lu, and Jianzhuang Liu.
\newblock Neighbor2neighbor: Self-supervised denoising from single noisy
  images.
\newblock In {\em CVPR}, 2021.

\bibitem{Adam}
Diederik~P Kingma and Jimmy Ba.
\newblock Adam: A method for stochastic optimization.
\newblock In {\em ICLR}, 2015.

\bibitem{N2V}
Alexander Krull, Tim-Oliver Buchholz, and Florian Jug.
\newblock Noise2void-learning denoising from single noisy images.
\newblock In {\em CVPR}, 2019.

\bibitem{ks2022content}
Green~Rosh KS, Nikhil Krishnan, BH~Pawan Prasad, and Sachin Lomte.
\newblock Content preserving scale space network for fast image restoration
  from noisy-blurry pairs.
\newblock In {\em ICASSP}, 2022.

\bibitem{Laine19}
Samuli Laine, Tero Karras, Jaakko Lehtinen, and Timo Aila.
\newblock High-quality self-supervised deep image denoising.
\newblock In {\em NeurIPS}, 2019.

\bibitem{AP-BSN}
Wooseok Lee, Sanghyun Son, and Kyoung~Mu Lee.
\newblock Ap-bsn: Self-supervised denoising for real-world images via
  asymmetric pd and blind-spot network.
\newblock In {\em CVPR}, 2022.

\bibitem{Noise2Noise}
Jaakko Lehtinen, Jacob Munkberg, Jon Hasselgren, Samuli Laine, Tero Karras,
  Miika Aittala, and Timo Aila.
\newblock Noise2noise: Learning image restoration without clean data.
\newblock In {\em ICML}, 2018.

\bibitem{MWCNN}
Pengju Liu, Hongzhi Zhang, Kai Zhang, Liang Lin, and Wangmeng Zuo.
\newblock Multi-level wavelet-cnn for image restoration.
\newblock In {\em CVPR}, 2018.

\bibitem{NRQM}
Chao Ma, Chih-Yuan Yang, Xiaokang Yang, and Ming-Hsuan Yang.
\newblock Learning a no-reference quality metric for single-image
  super-resolution.
\newblock {\em Computer Vision and Image Understanding}, 2017.

\bibitem{michaeli2014blind}
Tomer Michaeli and Michal Irani.
\newblock Blind deblurring using internal patch recurrence.
\newblock In {\em ECCV}, 2014.

\bibitem{KPN}
Ben Mildenhall, Jonathan~T Barron, Jiawen Chen, Dillon Sharlet, Ren Ng, and
  Robert Carroll.
\newblock Burst denoising with kernel prediction networks.
\newblock In {\em CVPR}, 2018.

\bibitem{NIQE}
Anish Mittal, Rajiv Soundararajan, and Alan~C Bovik.
\newblock Making a “completely blind” image quality analyzer.
\newblock {\em IEEE Signal Processing Letters}, 2012.

\bibitem{LSD2}
Janne Mustaniemi, Juho Kannala, Jiri Matas, Simo S{\"a}rkk{\"a}, and Janne
  Heikkil{\"a}.
\newblock Lsd$_2$--joint denoising and deblurring of short and long exposure
  images with cnns.
\newblock In {\em BMVC}, 2020.

\bibitem{nah2019ntire}
Seungjun Nah, Sungyong Baik, Seokil Hong, Gyeongsik Moon, Sanghyun Son, Radu
  Timofte, and Kyoung Mu~Lee.
\newblock Ntire 2019 challenge on video deblurring and super-resolution:
  Dataset and study.
\newblock In {\em CVPR Workshops}, 2019.

\bibitem{DeepDeblur}
Seungjun Nah, Tae~Hyun Kim, and Kyoung~Mu Lee.
\newblock Deep multi-scale convolutional neural network for dynamic scene
  deblurring.
\newblock In {\em CVPR}, 2017.

\bibitem{CVF-SID}
Reyhaneh Neshatavar, Mohsen Yavartanoo, Sanghyun Son, and Kyoung~Mu Lee.
\newblock Cvf-sid: Cyclic multi-variate function for self-supervised image
  denoising by disentangling noise from image.
\newblock In {\em CVPR}, 2022.

\bibitem{R2R}
Tongyao Pang, Huan Zheng, Yuhui Quan, and Hui Ji.
\newblock Recorrupted-to-recorrupted: Unsupervised deep learning for image
  denoising.
\newblock In {\em CVPR}, 2021.

\bibitem{PyTorch}
Adam Paszke, Sam Gross, Francisco Massa, Adam Lerer, James Bradbury, Gregory
  Chanan, Trevor Killeen, Zeming Lin, Natalia Gimelshein, Luca Antiga, Alban
  Desmaison, Andreas Kopf, Edward Yang, Zachary DeVito, Martin Raison, Alykhan
  Tejani, Sasank Chilamkurthy, Benoit Steiner, Lu Fang, Junjie Bai, and Soumith
  Chintala.
\newblock Pytorch: An imperative style, high-performance deep learning library.
\newblock In {\em NeurIPS}, 2019.

\bibitem{DND}
Tobias Plotz and Stefan Roth.
\newblock Benchmarking denoising algorithms with real photographs.
\newblock In {\em CVPR}, 2017.

\bibitem{SelfDeblur}
Dongwei Ren, Kai Zhang, Qilong Wang, Qinghua Hu, and Wangmeng Zuo.
\newblock Neural blind deconvolution using deep priors.
\newblock In {\em CVPR}, 2020.

\bibitem{ren2018deep}
Wenqi Ren, Jiawei Zhang, Lin Ma, Jinshan Pan, Xiaochun Cao, Wangmeng Zuo, Wei
  Liu, and Ming-Hsuan Yang.
\newblock Deep non-blind deconvolution via generalized low-rank approximation.
\newblock In {\em NeurIPS}, 2018.

\bibitem{RealBlur}
Jaesung Rim, Haeyun Lee, Jucheol Won, and Sunghyun Cho.
\newblock Real-world blur dataset for learning and benchmarking deblurring
  algorithms.
\newblock In {\em ECCV}, 2020.

\bibitem{UNet}
Olaf Ronneberger, Philipp Fischer, and Thomas Brox.
\newblock U-net: Convolutional networks for biomedical image segmentation.
\newblock In {\em MICCAI}, 2015.

\bibitem{schuler2015learning}
Christian~J Schuler, Michael Hirsch, Stefan Harmeling, and Bernhard
  Sch{\"o}lkopf.
\newblock Learning to deblur.
\newblock {\em IEEE TPAMI}, 2015.

\bibitem{shan2008high}
Qi Shan, Jiaya Jia, and Aseem Agarwala.
\newblock High-quality motion deblurring from a single image.
\newblock {\em ACM TOG}, 2008.

\bibitem{HIDE}
Ziyi Shen, Wenguan Wang, Xiankai Lu, Jianbing Shen, Haibin Ling, Tingfa Xu, and
  Ling Shao.
\newblock Human-aware motion deblurring.
\newblock In {\em ICCV}, 2019.

\bibitem{soltanayev2018training}
Shakarim Soltanayev and Se~Young Chun.
\newblock Training deep learning based denoisers without ground truth data.
\newblock In {\em NeurIPS}, 2018.

\bibitem{PWC-Net}
Deqing Sun, Xiaodong Yang, Ming-Yu Liu, and Jan Kautz.
\newblock Pwc-net: Cnns for optical flow using pyramid, warping, and cost
  volume.
\newblock In {\em CVPR}, 2018.

\bibitem{tao2018scale}
Xin Tao, Hongyun Gao, Xiaoyong Shen, Jue Wang, and Jiaya Jia.
\newblock Scale-recurrent network for deep image deblurring.
\newblock In {\em CVPR}, 2018.

\bibitem{DIP}
Dmitry Ulyanov, Andrea Vedaldi, and Victor Lempitsky.
\newblock Deep image prior.
\newblock In {\em CVPR}, 2018.

\bibitem{SSIM}
Zhou Wang, Alan~C Bovik, Hamid~R Sheikh, and Eero~P Simoncelli.
\newblock Image quality assessment: from error visibility to structural
  similarity.
\newblock {\em IEEE TIP}, 2004.

\bibitem{Blind2Unblind}
Zejin Wang, Jiazheng Liu, Guoqing Li, and Hua Han.
\newblock Blind2unblind: Self-supervised image denoising with visible blind
  spots.
\newblock In {\em CVPR}, 2022.

\bibitem{wieschollek2017learning}
Patrick Wieschollek, Michael Hirsch, Bernhard Scholkopf, and Hendrik Lensch.
\newblock Learning blind motion deblurring.
\newblock In {\em ICCV}, 2017.

\bibitem{DBSN}
Xiaohe Wu, Ming Liu, Yue Cao, Dongwei Ren, and Wangmeng Zuo.
\newblock Unpaired learning of deep image denoising.
\newblock In {\em ECCV}, 2020.

\bibitem{BPN}
Zhihao Xia, Federico Perazzi, Micha{\"e}l Gharbi, Kalyan Sunkavalli, and Ayan
  Chakrabarti.
\newblock Basis prediction networks for effective burst denoising with large
  kernels.
\newblock In {\em CVPR}, 2020.

\bibitem{xu2013unnatural}
Li Xu, Shicheng Zheng, and Jiaya Jia.
\newblock Unnatural l0 sparse representation for natural image deblurring.
\newblock In {\em CVPR}, 2013.

\bibitem{SIGGRAPH-2007}
Lu Yuan, Jian Sun, Long Quan, and Heung-Yeung Shum.
\newblock Image deblurring with blurred/noisy image pairs.
\newblock In {\em SIGGRAPH}, 2007.

\bibitem{DANet}
Zongsheng Yue, Qian Zhao, Lei Zhang, and Deyu Meng.
\newblock Dual adversarial network: Toward real-world noise removal and noise
  generation.
\newblock In {\em ECCV}, 2020.

\bibitem{Restormer}
Syed~Waqas Zamir, Aditya Arora, Salman Khan, Munawar Hayat, Fahad~Shahbaz Khan,
  and Ming-Hsuan Yang.
\newblock Restormer: Efficient transformer for high-resolution image
  restoration.
\newblock In {\em CVPR}, 2022.

\bibitem{MIRNet}
Syed~Waqas Zamir, Aditya Arora, Salman Khan, Munawar Hayat, Fahad~Shahbaz Khan,
  Ming-Hsuan Yang, and Ling Shao.
\newblock Learning enriched features for real image restoration and
  enhancement.
\newblock In {\em ECCV}, 2020.

\bibitem{MPRNet}
Syed~Waqas Zamir, Aditya Arora, Salman Khan, Munawar Hayat, Fahad~Shahbaz Khan,
  Ming-Hsuan Yang, and Ling Shao.
\newblock Multi-stage progressive image restoration.
\newblock In {\em CVPR}, 2021.

\bibitem{zhang2018dynamic}
Jiawei Zhang, Jinshan Pan, Jimmy Ren, Yibing Song, Linchao Bao, Rynson~WH Lau,
  and Ming-Hsuan Yang.
\newblock Dynamic scene deblurring using spatially variant recurrent neural
  networks.
\newblock In {\em CVPR}, 2018.

\bibitem{DnCNN}
Kai Zhang, Wangmeng Zuo, Yunjin Chen, Deyu Meng, and Lei Zhang.
\newblock Beyond a gaussian denoiser: Residual learning of deep cnn for image
  denoising.
\newblock {\em IEEE TIP}, 2017.

\bibitem{FFDNet}
Kai Zhang, Wangmeng Zuo, and Lei Zhang.
\newblock Ffdnet: Toward a fast and flexible solution for cnn-based image
  denoising.
\newblock {\em IEEE TIP}, 2018.

\bibitem{LPIPS}
Richard Zhang, Phillip Isola, Alexei~A Efros, Eli Shechtman, and Oliver Wang.
\newblock The unreasonable effectiveness of deep features as a perceptual
  metric.
\newblock In {\em CVPR}, 2018.

\bibitem{IDR}
Yi Zhang, Dasong Li, Ka~Lung Law, Xiaogang Wang, Hongwei Qin, and Hongsheng Li.
\newblock Idr: Self-supervised image denoising via iterative data refinement.
\newblock In {\em CVPR}, 2022.

\bibitem{zhou2019davanet}
Shangchen Zhou, Jiawei Zhang, Wangmeng Zuo, Haozhe Xie, Jinshan Pan, and
  Jimmy~S Ren.
\newblock Davanet: Stereo deblurring with view aggregation.
\newblock In {\em CVPR}, 2019.

\bibitem{zhussip2019extending}
Magauiya Zhussip, Shakarim Soltanayev, and Se~Young Chun.
\newblock Extending stein's unbiased risk estimator to train deep denoisers
  with correlated pairs of noisy images.
\newblock In {\em NeurIPS}, 2019.

\end{thebibliography}
}

\end{document}